\documentclass[journal]{IEEEtran}
\usepackage[latin1]{inputenc}
\usepackage{graphicx}

\usepackage{algorithm}
\usepackage{algorithmic}
\usepackage{color}
 \usepackage{amsmath,amsfonts,amssymb}
\begin{document}

\title{Conceptualization of seeded region growing by pixels aggregation. Part 3: a wide range of algorithms.}
\author{Vincent Tariel}
\maketitle
\begin{abstract}
In the two previous papers of this serie, we have created a library, called Population, dedicated to seeded region growing by pixels aggregation and we have proposed different growing processes to get a partition with or without a boundary region to divide the other regions or to get a partition invariant about the seeded region initialisation order. Using this work, we implement some algorithms belonging to the field of SRGPA  using this library and these growing processes.
\end{abstract}

\IEEEpeerreviewmaketitle

\section{Introduction}
Many fields in computer science, stereovision\cite{KANADE1994}, mathematical morphology\cite{Serra1982},
use algorithm which principle is Seeded Region Growing by Pixels Aggregation (SRGPA). This method consists in initializing each region with a seed, then processing pixels aggregation on regions, iterating this aggregation until getting a nilpotence \cite{ADAMS1994}\cite{Hojjatoleslami1998}. The general purpose of this field is to define a metric divided into two distinct categories \cite{Ballard1982}: the region feature like the tint \cite{ADAMS1994} and region boundary discontinuity\cite{Beucher1979}.\\
In this article, the aim is not to do an overview of the algorithms using SRGPA but to prove that the framework introduced in the two previous articles\cite{Tariel2008b}\cite{Tariel2008c} is generic. Some algorithms using SRGPA are implemented thanks to the library Population:
\begin{itemize}
\item voronoï tessellation, regional minima, domain to clusters,
\item distance function, watershed transformation and geodesic reconstruction.
\end{itemize}
The first enhancement is the easiness to implement these algorithms  using the objects of the library Population. The second enhancement is the algorithms efficiency. All these algorithms have been applied on 3D image with a size equal to 700*700*700=0.35 Giga pixels. The running time is always less than 3 hours with an Intel(R) Xeon(R) CPU 3.00GH. This is due to
\begin{enumerate}
\item the library optimisation using the template metaprogramming\footnote{Template metaprogramming is a metaprogramming technique in which templates are used by a compiler to generate temporary source code, which is merged by the compiler with the rest of the source code and then compiled. The output of these templates include compile-time constants, data structures, and complete functions. The use of templates can be thought of as compile-time execution.}\cite{Alexandrescu2001}: all algorithms using this library will benefit from this optimization,
\item the procedure of actualization of the zones of influence described in the previous article\cite{Tariel2008b}.
\end{enumerate}
In this article, the notations are:
\begin{itemize}
\item let $E$ be a discrete space\footnote{ The space $E$, is a n-dimensional discrete space $\mathbb{Z}^n$, consisting of lattice points whose coordinates are all integers in a three-dimensional Euclidean space $\mathbb{R}^n$. The elements of a n-dimensional image array are called points.},
\item let $\Omega$ be a domain of $E$  and $I$ its characteristic function such as $\Omega=\{\forall x\in E: I(x)\neq 0\}$,
\item let $f$ be a grey-level image, an application of $E$ to  $\mathbb{Z}$,
\item let $V$ be a neighborhood function (an elementary structuring element).
\end{itemize}
 In the appendice~\ref{ap:dist}, the definition of the distance is given. The article understanding depends on the comprehension of the previous articles of this serie. A summary is done in the appendice~\ref{ap:sum}.\\
The outline of the rest of the paper is as follows: in Sec.~II, we present the algorithms using only one queue in the system of queue (SQ), in Sec.~III we present the algorithms using more than one queue, in Sec.~IV, we make concluding remarks. 

\section{One queue}
In this section, we will present some algorithms using a single queue during the growing process.

\subsection{Simulated Vorono\"i tessellation}
\begin{figure}
\begin{center}
\includegraphics[width=1.6cm]{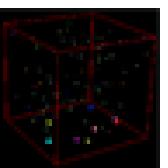}
\includegraphics[width=1.6cm]{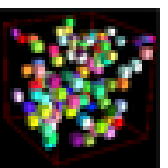}
\includegraphics[width=1.6cm]{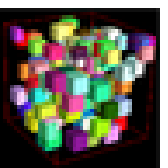}
\includegraphics[width=1.6cm]{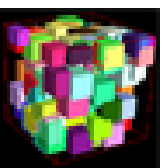}
\includegraphics[width=1.6cm]{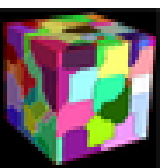}
\includegraphics[width=1.6cm]{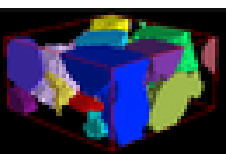}
\includegraphics[width=1.6cm]{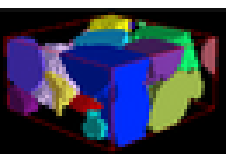}
\includegraphics[width=1.6cm]{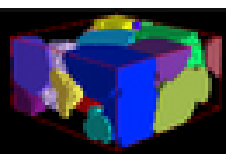}
\includegraphics[width=1.6cm]{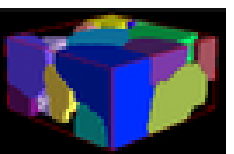}
\includegraphics[width=1.6cm]{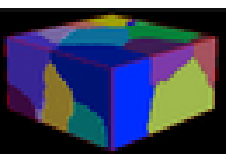}
\includegraphics[width=1.6cm]{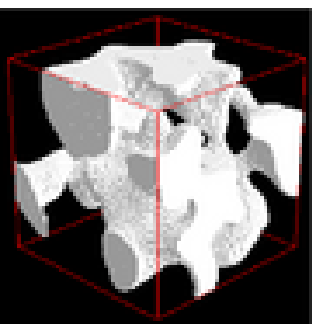}
\includegraphics[width=1.6cm]{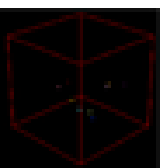}
\includegraphics[width=1.6cm]{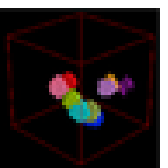}
\includegraphics[width=1.6cm]{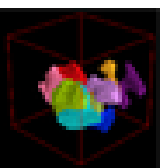}
\includegraphics[width=1.6cm]{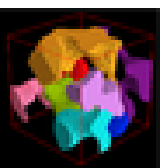}
\caption{Each serie serie is the Vorono\"i-growing process. For the first serie, the seeds are drawn from the Poisson point, \textit{A video is available at http://pmc.polytechnique.fr/$\sim$vta/geodesic$\_$invariant$\_$cube.mpeg}.  For the second serie, each seed is a set of points, \textit{A video is available at http://pmc.polytechnique.fr/$\sim$vta/geodesic$\_$invariant$\_$cube$\_$grain.mpeg}. For the last serie, the growing process is restricted by the first figure,  \textit{A video is available at http://pmc.polytechnique.fr/$\sim$vta/geodesic.mpeg}.}
\label{fig:vor}
\end{center} 
\end{figure}

Consider $\Phi$ a Poisson point process in a metric space $M$. The cells
\[
C(x) = \{y \in M; d(y - x) \leq d(y - x), x' \in \Phi\}, x\in \Phi,
\]
constitute the so-called Poisson-Vorono\"i tessellation of $M$. Presented by Gilbert in 1962 \cite{GILBERT1962}, this statistical model is appropriate for random crystal growth. In the discrete space $E$, the implementation for a distance associated to norm 1 or $\infty$\footnote{For the Euclidian distance, see~\cite{Vincent1991}. }\cite{Schmitt1989} is done using the library Population.\\
Starting form the affectation of each region with a seed (a point of Poisson point process), an isotopic growing process at constant velocity is operated. The ordering attribute function is $\delta(x,i)=0$.  The growing process is (see algorithm~\ref{alg1} and figure~\ref{fig:vor}):
\begin{itemize}	
\item initialization of the regions/ZI by the seeds
\item select the queue number 0
\item while the selected queue is not empty
\begin{itemize}
\item extract $(y,i)$ from the selected queue
\item "Growth on $x$ of the region $i$"
\end{itemize}
\item return regions
\end{itemize}
The quote "growth on $x$ of the region i" means that there are different kinds of growing process introduced in the previous article~\cite{Tariel2008c}. Here, the growing process is done without a boundary region to divide the other regions. In the algorithm~\ref{alg1}, the growing process leading to a final partition invariant about the seeded region initialisation order is used. To prove that this growing process gives a correct Poisson-Vorono\"i tessellation of $E$, this property is used:\\ 
\[
\forall x, y \in E : d(x,y)=\min_{z\in \overline{V(x)}}((d(x,z)+1)
\]
The generation of a Poisson point process is done using the Boost software. This implementation is not restricted to the Poisson-Vorono\"i tessellation since:
\begin{itemize}
\item each seed can be a domain of $E$ (second serie in the figure~\ref{fig:vor}),
\item the growing process can be restricted to a domain $\Omega=\{\forall x\in E: I(x)\neq 0\}$ if the ordering attribute function is: $\delta(x,i)=0 \mbox{ if } I(x)\neq 0, OUT \mbox{ else }$ (third serie in the figure~\ref{fig:vor}).
\end{itemize}
 \begin{algorithm}[h!tp]
\caption{Geodesic dilatation with an invariant boundary}
\label{alg1}
\algsetup{indent=1em}
\begin{algorithmic}[20]
 \REQUIRE $S$ , $V$ \textit{//The binary image, the seeds, the neighborhood}
\STATE \textbf{\textit{// initialization}}
\STATE System$\_$Queue s$\_$q( $\delta(x,i)=0$, FIFO, 1); \textit{//A single FIFO queue}
\STATE Population p (s$\_$q); \textit{//create the object Population}
\STATE Tribe passive($V=\emptyset$);
\STATE int ref$\_$boundary   = p.growth$\_$tribe(passive);
\STATE \textit{//create a boundary region/ZI, $(X^t_b,Z^t_b)$ such as $Z^t_i=\emptyset$}
\STATE Restricted $N$=$\mathbb{N}$; 
\STATE Tribe active(V, N);
\FORALL{$\forall s_i\in S$} 
\STATE int ref$\_$tr   = p.growth$\_$tribe(actif); \textit{//create a region/ZI, $(X^t_i,Z^t_i)$ such as $Z^t_i=(X_{i}^t \oplus V)\setminus (\bigcup\limits_{j \in \mathbb{N}} X_{j})$}
\STATE  p.growth($s_i$, ref$\_$tr ); 
\ENDFOR
\STATE \textbf{\textit{//the growing process}}
\STATE  s$\_$q.select$\_$queue(0); \textit{//select the single FIFO queue.}
\WHILE{s$\_$q.empty()==false}
\STATE  $(x,i)=s$\_$q$.pop();
\IF{pop.Z()[x].size()$\geq$2}
\STATE  p.growth(x, ref$\_$boundary); \textit{//growth of the boundary region}
\ELSE
\STATE  p.growth(x, i ); \textit{//simple growth}
\ENDIF
\ENDWHILE
\RETURN p.X();
\end{algorithmic}
 \end{algorithm}
\subsection{Domain to clusters}
\begin{figure}[h]
    \begin{center}
\includegraphics[width=2.6cm]{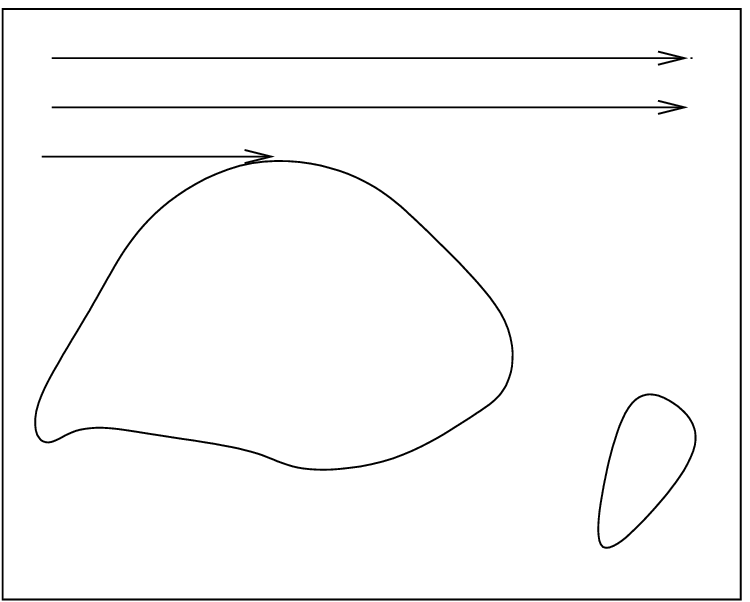}
\includegraphics[width=2.6cm]{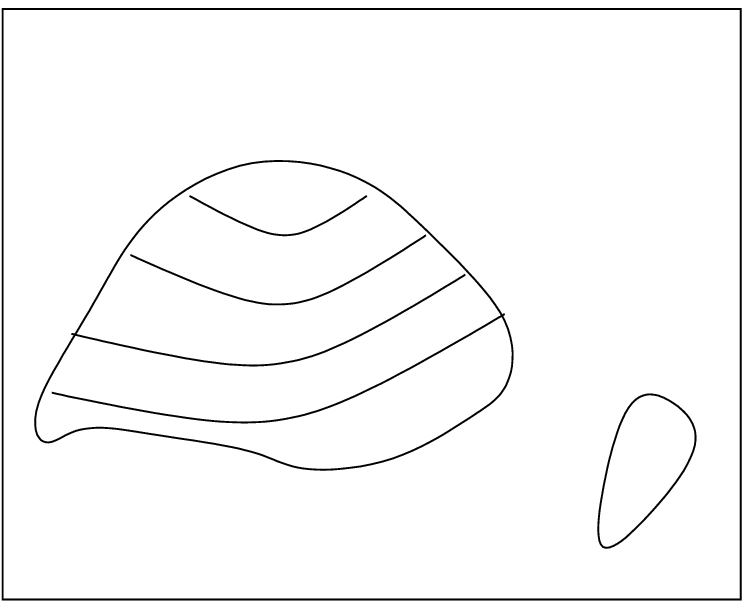}
\includegraphics[width=2.6cm]{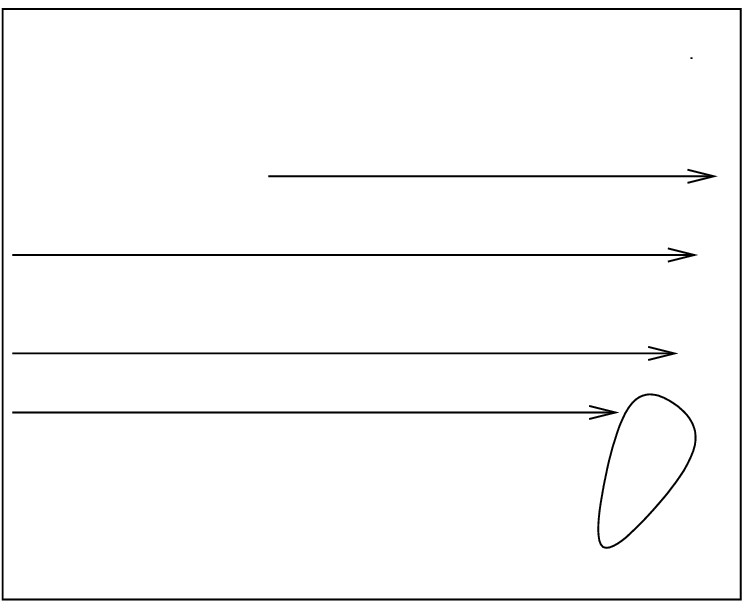}
\caption{Left  image: scan the image until $I(x)\neq 0$, middle  image: growing region starting form   $x$ ($I(x)\neq 0$) such as at each growth the characteristic function of $\Omega$ is modified $I(x)\neq 0\rightarrow I(x)= 0$, right  image: at the end of the growing region, the connected  component  has been extracted and removed from $\Omega$ and the scanning continues until $I(x)\neq 0$.} 
        \label{cluster}
\end{center}
\end{figure}

\begin{figure}[h]
    \begin{center}
\includegraphics[width=2cm]{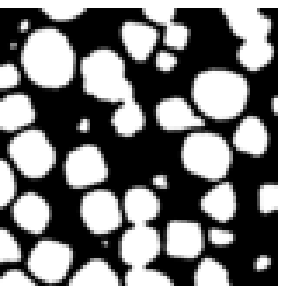}
\includegraphics[width=2cm]{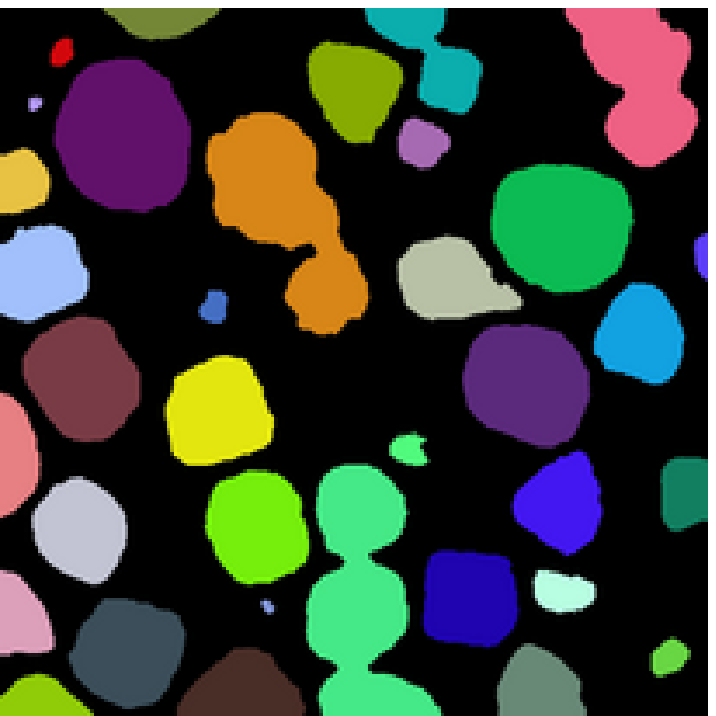}
\includegraphics[width=2cm]{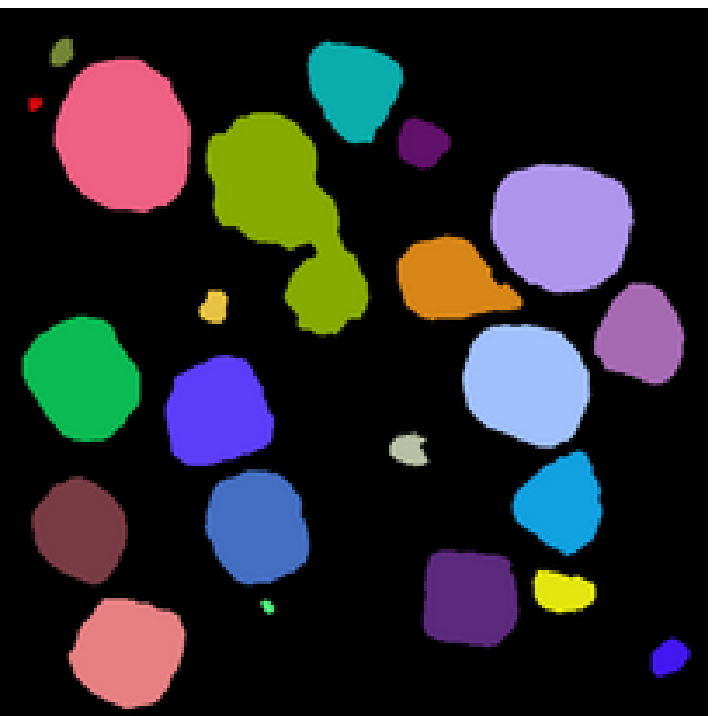}
\includegraphics[width=2cm]{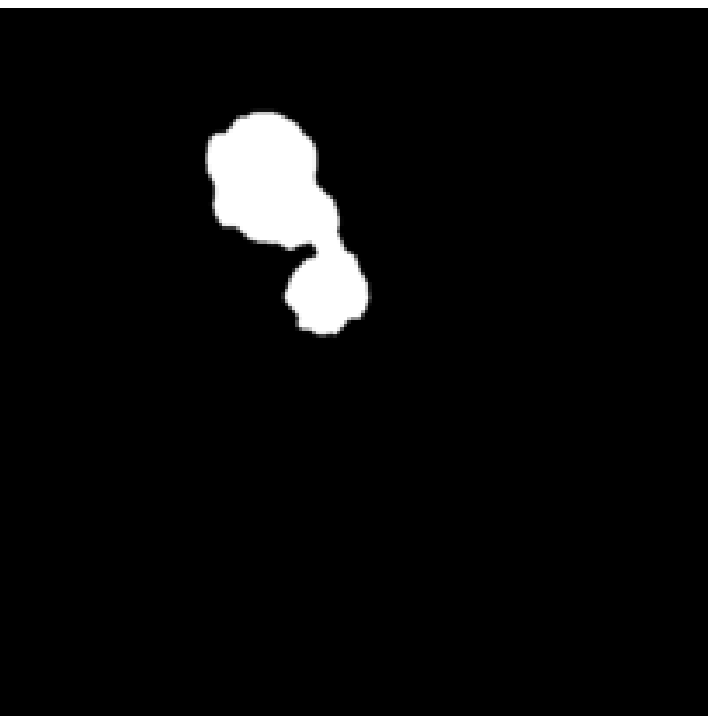}
\includegraphics[width=4cm]{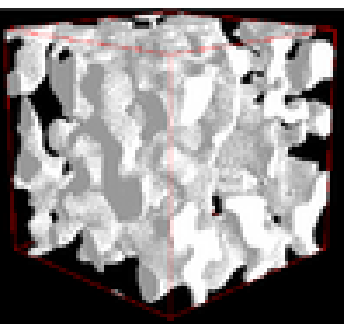}
\includegraphics[width=4cm]{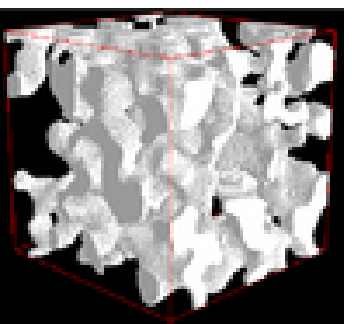}
\caption{Left upper image: the initial binary image, left middle upper image: extraction of the connex components, right middle upper image: the connected components touching the boundary of the image are removed, right upper image: the max cluster of the previous image, left bottom image: the initial image, right bottom image: maximum cluster of percolation after the selection of the component whose the area is maximum in the connected components.} 
        \label{clusterapp}
\end{center}
\end{figure}

Let $\Omega$ be a domain of $E$ and let $\mathcal{C}^\Omega$ be the set of continuous application from $[0,1]$ to $\Omega$.\\
$(c_i)_{0\leq i \leq n}$ is the clusters decomposition of $\Omega$  if:
\begin{eqnarray*}
&&\cup_{0\leq i \leq n}c_i=\Omega\\
\forall i \in (0,\ldots,n) \forall x,y \in (c_i,c_i)& \exists \gamma \in \mathcal{C}^\Omega& \gamma(0)=x \wedge \gamma(1)=y\\
\forall i \neq j \forall x,y \in (c_i,c_j) &\nexists \gamma \in \mathcal{C}^\Omega &\gamma(0)=x \wedge \gamma(1)=y
\end{eqnarray*}
The second line means that all points belonging to the same connected component are linked and the third line means that two points belonging to different connected components are not linked. This extraction gives information about the critical percolation concentration, percolation probabilities, and cluster size distributions\cite{HOSHEN1976}. Using the library Population, a algorithm is defined to extract the set of connected components. The principle is: when a connected component is touched, this connected component is removed from $\Omega$ using a growing process (see algorithm~\ref{alg:cluster} and figure~\ref{cluster}):
\begin{itemize}
\item scan the image ($\forall x\in E$)
\begin{itemize}
\item if $I(x)\neq 0$
\begin{itemize}
\item create a region/ZI initialised by the seed $\{x\}$
\item select the queue number 0
\item while the selected queue is not empty
\begin{itemize}
\item extract $(y,i)$ from the selected queue
\item growth of the region $i$ on $y$
\item $I(y)=0$
\end{itemize}
\end{itemize}
\end{itemize}
\item return regions
\end{itemize}
Using this extraction, it is possible (see figure~\ref{clusterapp}):
\begin{itemize}
\item to remove all the connected components touching the boundary,
\item to fill the hole\footnote{To file the hole, the porcedure is
\begin{enumerate}
\item inversion of the initial image,
\item extraction of connected components,
\item removing the connected components no touching the image boundary,
\item binarization and inversion of this last image.
\end{enumerate}},
\item to keep only the cluster which area is maximum,
% \item  to c   the geometrical tortuosity\cite{Delarue2001}
\end{itemize}
 \begin{algorithm}[h!tp]
\caption{Domain to clusters}
\label{alg:cluster}
\algsetup{indent=1em}
\begin{algorithmic}[20]
 \REQUIRE $I$,  $V$ \textit{//The binary image, the neighborhood}
\STATE \textbf{\textit{// initialization}}
\STATE System$\_$Queue s$\_$q( $\delta(x,i)=0\mbox{ if }I(x)\neq 0, OUT\mbox{ else}$, FIFO, 1); \textit{//A single FIFO queue such as if $I(x)=0$ then $(x,i)$ is not pushed in the SQ.}
\STATE Population p (s$\_$q); \textit{//create the object Population}
\STATE Restricted $N=\mathbb{N}$;
\STATE Tribe active(V, N);
\STATE  \textbf{\textit{//Scan the image}}
\FORALL{$\forall x \in E$}
\STATE  \textit{//Test if a connex component is touched}
\IF{$I(x)\neq 0$} 
\STATE int ref$\_$tr   = p.growth$\_$tribe(actif); \textit{//create a region/ZI, $(X^t_i,Z^t_i)$ such as $Z^t_i=(X_{i}^t \oplus V)\setminus (\bigcup\limits_{j \in \mathbb{N}} X_{j})$}
\STATE  p.growth(x, ref$\_$tr ); 
\STATE \textbf{\textit{//the growing process}}
\STATE  s$\_$q.select$\_$queue(0); \textit{//select the single FIFO queue.}
\WHILE{s$\_$q.empty()==false}
\STATE  $(y,i)=s$\_$q$.pop();
\STATE p.growth($y, i$ );
\STATE $I(y)= 0$;  
\ENDWHILE
\ENDIF
\ENDFOR
\RETURN p.X();
\end{algorithmic}
 \end{algorithm}
\subsection{Regional minima}
\begin{figure}[h]
    \begin{center}
\includegraphics[width=6cm]{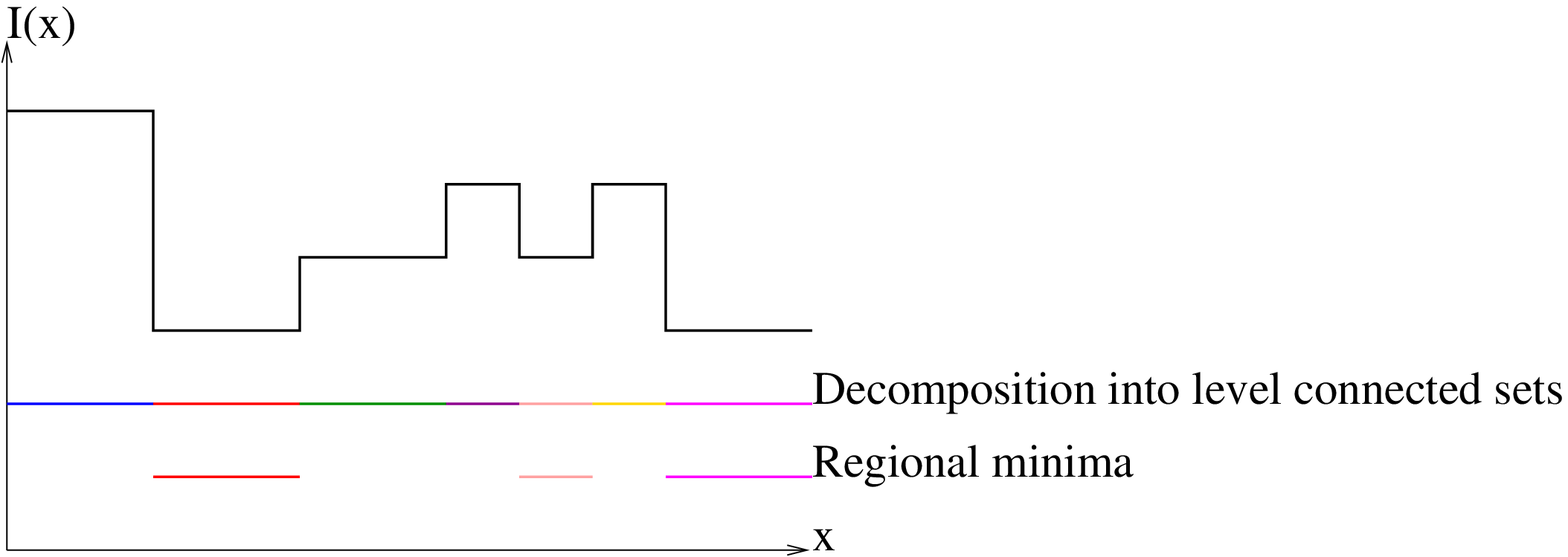}
\includegraphics[width=3cm]{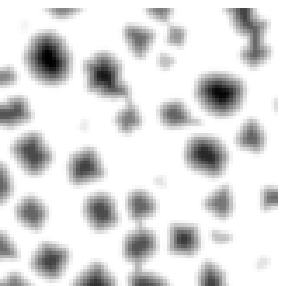}
\includegraphics[width=3cm]{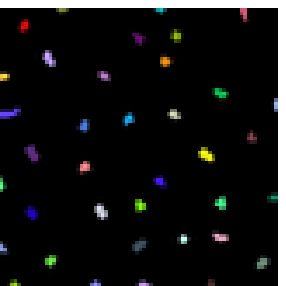}
\caption{Upper image: principle of minima; bottom images: on the left, a grey-level image, on the right the regional minima of this image} 
        \label{minima}
\end{center}
\end{figure}

Let $\mathcal{C}^E_{x,y}$ be the set of continuous application from $[0,1]$ to $E$ such as the two extremities are equal to $x$ and $y$ ($\forall \gamma \in \mathcal{C}_{x,y}: \gamma(0)=x \wedge\gamma(1)=y$).\\
$S=(s_i)_{0\leq i \leq n}$ is the decomposition of $(E,I)$ in level connected  sets if:
\begin{eqnarray*}
&\bigcup\limits_{0\leq i \leq n}s_i=E\\
\forall i \in (0,\ldots,n) \forall x,y \in s_i&   I(x)=I(y)\\
\forall i\neq j \forall \gamma \in  \mathcal{C}^E_{x,y} \exists t \in [0,1] &  (I(x)\neq I(\gamma(t))) \vee (I(y)\neq I(\gamma(t)))
\end{eqnarray*}
If $f$ is seen as a topographic surface, the second line means that the level is the same in each point belonging to $s_i$ and the third line means that all paths between two points belonging to different elements of $S$ do not have a constant level.\\ 
In this decomposition, an element $s$ of $S$ is a regional minimum if:
\begin{eqnarray*}
\forall (x, y) \in (s, (s\oplus V) \setminus s) &   I(x)<I(y)
\end{eqnarray*}
The level of the points belonging to the outer boundary of $s$ is greater than the level of the points belonging to $s$ (see figure~\ref{minima}). Using the library Population, a growing procedure is defined to extract the regional minima. This growing procedure consists to scan the image ($\forall x\in E$). At each time, there is not yet a region on $x$ ($pop.X()[x].empty()==true$) to start the growing region initialized by the seed equal to $\{x\}$. Let $level=I(x)$ be the level of the growing region. The ordering attribute function is defined as:
\[
\delta(x,i)=0 \mbox{ if }I(x)\leq level, OUT \mbox{ else}
\]
 For this algorithm, the ZI is defined as: $Z^t_i=(X_{i}^t \oplus V)\setminus X_{i}^t$ because the ZI is localized on the outter boundary region even if there are still some region to check the condition: $\forall (x, y) \in (s, (s\oplus V) \setminus s) :  I(x)<I(y)$. The growing process is (see algorithm~\ref{alg:minima} and figure~\ref{minima}):
\begin{itemize}
\item to scan the image ($\forall x\in E$)
\begin{itemize}
\item if $pop.X()[x].empty()==true$
\begin{itemize}
\item create a region/ZI initialised by the seed $\{x\}$
\item level = I(x) 
\item select the queue number 0
\item while the selected queue is not empty
\begin{itemize}
\item extract $(y,i)$ from the selected queue
\item if $I(y)==level$
\item then growth of the region $i$ on $y$
\item else
\item then this region/ZI is not a regional minimum
\end{itemize}
\end{itemize}
\end{itemize}
\item return regions that are regional minima
\end{itemize}

 \begin{algorithm}[h!tp]
\caption{Regional minima}
\label{alg:minima}
\algsetup{indent=1em}
\begin{algorithmic}[40]
 \REQUIRE $I$, $V$ \textit{//The grey-value image, the neighborhood}
\STATE \textbf{\textit{// initialization}}
\STATE int $level$;
\STATE System$\_$Queue s$\_$q( $\delta(x,i)=0\mbox{ if }I(x)\leq level, OUT\mbox{ else}$, FIFO, 1); \textit{//A single FIFO queue}
\STATE Population p (s$\_$q); \textit{//create the object Population}
\STATE \fbox{Restricted $N=\{i\}$;}
\STATE Tribe active(V, N);
\STATE Set set; \textit{//Container: self-balancing binary search tree.}
\STATE \textbf{ \textit{//Scan the image}}
\FORALL{$\forall x \in E$}
\STATE  \textit{//Test if there is still a region on $x$}
\IF{$pop.X()[x].empty()==true$} 
\STATE $level = I(x)$
\STATE int ref$\_$tr   = p.growth$\_$tribe(actif); \textit{//create a region/ZI, $(X^t_i,Z^t_i)$ such as \fbox{$Z^t_i=(X_{i}^t \oplus V)\setminus X_{i}^t$}}
\STATE  p.growth(x, ref$\_$tr );
\STATE  bool regional$\_$minima=true; 
\STATE \textbf{\textit{//the growing process}}
\STATE  s$\_$q.select$\_$queue(0); \textit{//Select the single FIFO queue.}
\WHILE{s$\_$q.empty()==false}
\STATE  $(y,i)=s$\_$q$.pop();
\IF{$I(y)<level$}
\STATE regional$\_$minima=false;
\ELSE
\STATE p.growth($y, i$ );
\ENDIF
\ENDWHILE
\IF{regional$\_$minima==true}
\STATE set.insert(ref$\_$tr);
\ENDIF
\ENDIF
\ENDFOR
\RETURN (p.X(),set);
\end{algorithmic}
 \end{algorithm}
 \section{n queues}
In this section, we will present some algorithms such as a muti-queue is used during the growing process.
\subsection{Distance function: flip-flop queue}
\begin{figure}
\begin{center}
\includegraphics[width=3cm]{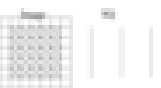}
\includegraphics[width=4cm]{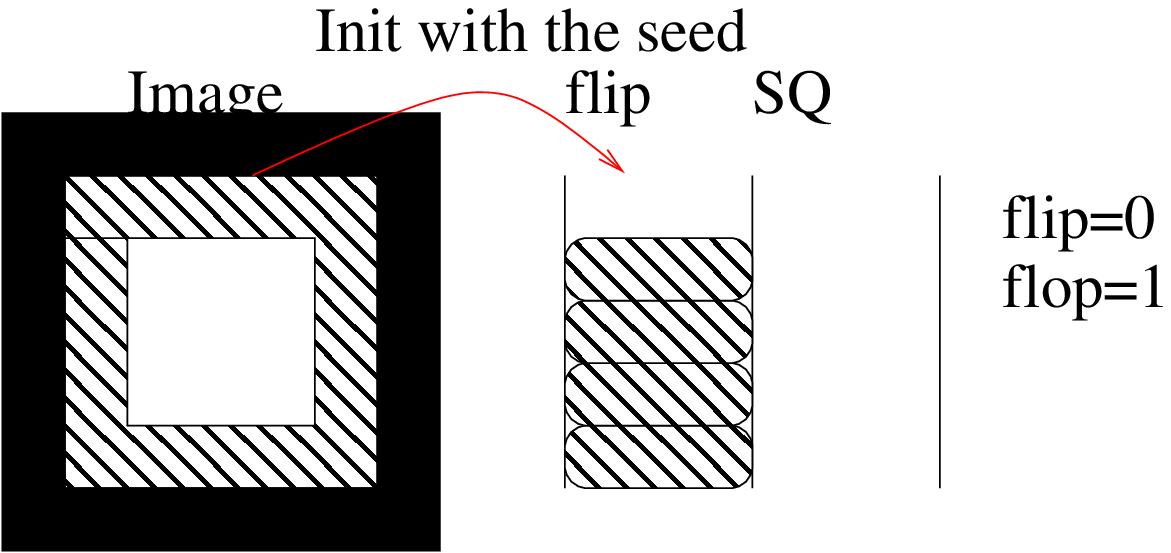}
\includegraphics[width=4cm]{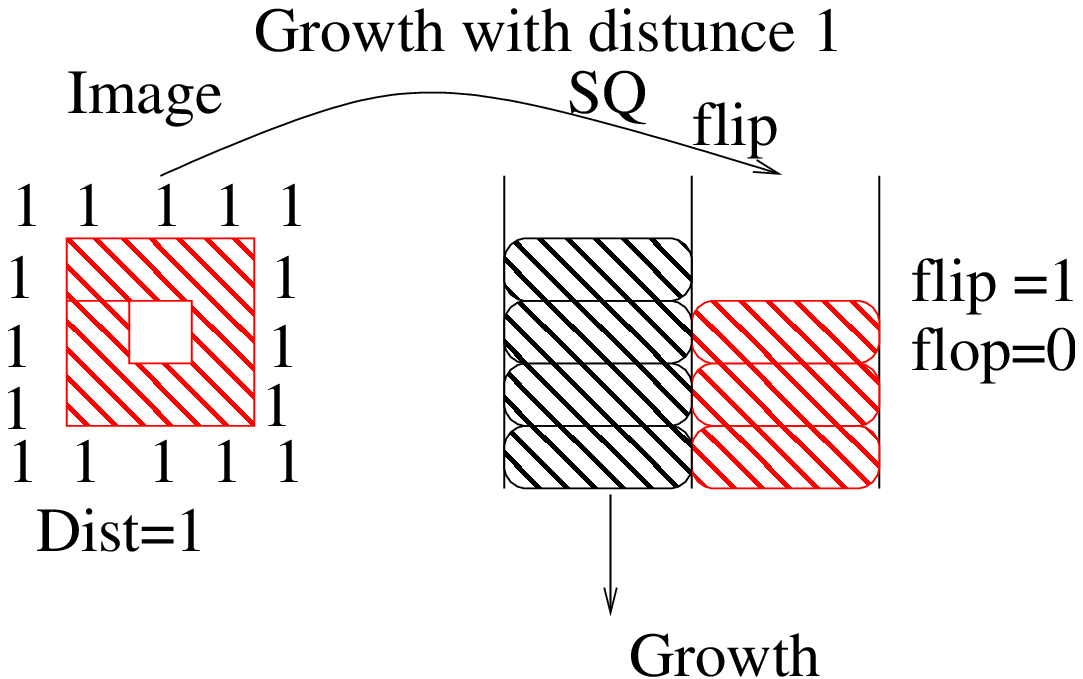}
\includegraphics[width=4cm]{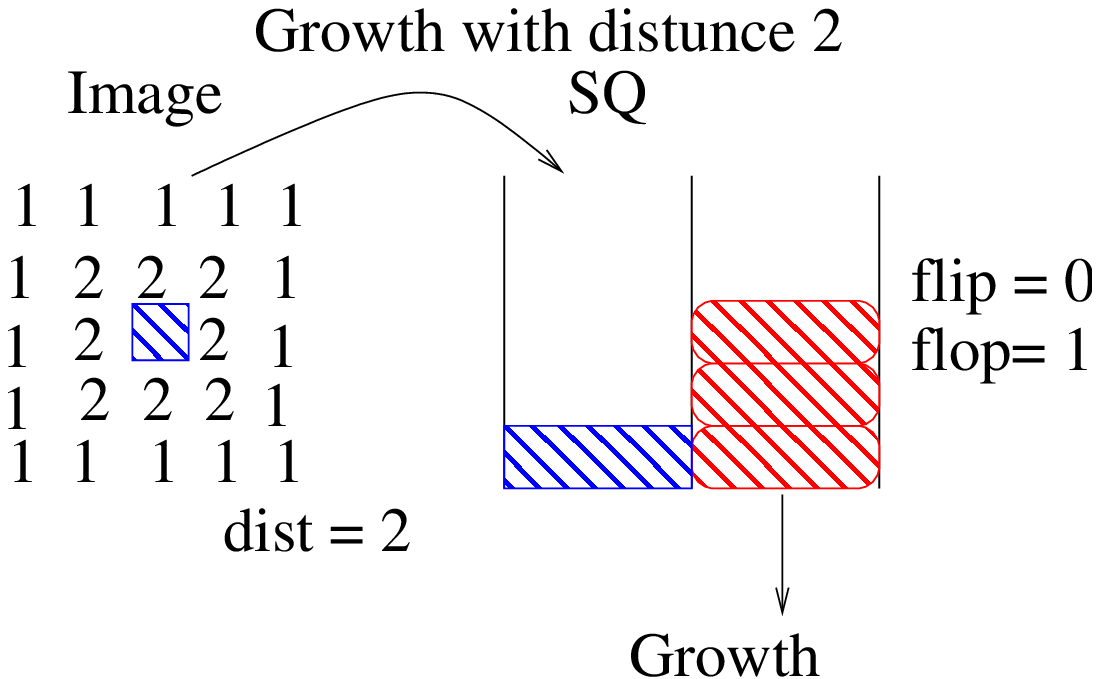}
\caption{The distance function}
\label{fig:dist}
\end{center} 
\end{figure}

For the voronoï tessellation, we impose only a growing process at a constant velocity with forgetting the distance between the seeds and the current position. Here, it is a growing process step by step where a step corresponds to a distance value. The output is a distance function.\\
% The distance function can be calculated in $E$ or a domain in $E$ but the algorithm is presented for the case of the whole space $E$ for the facility of the comprehension.\\
\paragraph{n queue implementation}
Let $d$ be the distance between a point and the seeds. The ordering attribute function is: $\delta(x,i) = d+1$ \footnote{The growing process is on the points which value is equal to $d$ and each couple $(x,i)$ is stored in the queue number $d+1$.}. The growing process is:
\begin{itemize}	
\item int d=0
\item initialization of the regions/ZI by the seeds
\item while the system of queues is not empty
\begin{itemize}
\item d = d + 1
\item select the queue number d
\item while the selected queue is not empty
\begin{itemize}
\item extract $(y,i)$ from the selected queue
\item growth on $x$ of the region $i$
\item dist[$x$]=$d$
\end{itemize}
\end{itemize}
\item return dist
\end{itemize}
The number of queues is equal to the maximum of the distance function. The problem of this implementation is that this number is unknown before the growing process. To overcome this problem, a solution is to use a flip-flop queue. 
\paragraph{flip-flop queue implementation}
In the last implementation, during the growing process, there are only two queues in the SQ not empty at the step $d$: the queue number $d$ where the couples are extracted and the number $d+1$ where the couples are stored. Using this property, the couples are now extracted from the queue number $flip$ and stored in the queue number $flop$. The ordering attribute function $\delta(x,i)$ is equal to $flop$.
The growing process becomes (see figure~\ref{fig:dist} and algorithm~\ref{alg:dist}):
\begin{itemize}	
\item int d=0
\item initialization of the regions/ZI by the seeds
\item while the system of queues is not empty
\begin{itemize}
\item d = d + 1
\item switch(flip,flop)
\item select the queue number flop
\item while the selected queue is not empty
\begin{itemize}
\item extract $(y,i)$ from the selected queue
\item growth on $x$ of the region $i$
\item dist[$x$]=$d$
\end{itemize}
\end{itemize}
\item return dist
\end{itemize}
 \begin{algorithm}[h!tp]
\caption{Distance function}
\label{alg:dist}
\algsetup{indent=1em}
\begin{algorithmic}[40]
 \REQUIRE $I$, $S$ , $V$ \textit{//the binary image, the seed, the neighborhood}
\STATE \textbf{\textit{// initialization}}
\STATE Image Img$\_$dist;
\STATE int flip=0, flop =1;
\STATE System$\_$Queue s$\_$q( $\delta(x,i)=flip \mbox{ if }I(x)\neq 0, OUT\mbox{ else}$, FIFO, 2); \textit{//two FIFO queues.}
\STATE Population p (s$\_$q); \textit{//create the object Population}
\STATE Restricted $N=\mathbb{N}$;
\FORALL{$\forall s_i\in S$ in the order $0,1\ldots$} 
\STATE int ref$\_$tr   = p.growth$\_$tribe(actif); \textit{//create a region/ZI, $(X^t_i,Z^t_i)$ such as $Z^t_i=(X_{i}^t \oplus V)\setminus (\bigcup\limits_{j \in \mathbb{N}} X_{j})$}
\STATE  p.growth($s_i$, ref$\_$tr ); 
\ENDFOR
\STATE int dist=0;
\STATE \textbf{\textit{//the growing process}}
\WHILE{s$\_$q.all$\_$empty()==false}
\STATE switch(flip,flop);
\STATE  s$\_$q.select$\_$queue(flop);
\STATE dist++;
\WHILE{s$\_$q.empty()==false}
\STATE  $(x,i)$=s$\_$q.pop();
\STATE p.growth($x, i$ );
\STATE Img$\_$dist($x$)= dist;  
\ENDWHILE
\ENDWHILE
\RETURN Img$\_$dist;
\end{algorithmic}
 \end{algorithm}
This algorithm is not limited to the distance function of $E$. The growing process can be restricted to a domain $\Omega=\{\forall x\in E: I(x)\neq 0\}$ if the ordering attribute function is: $\delta(x,i)=flop \mbox{ if } I(x)\neq 0, OUT \mbox{ else }$ (fourth serie in the figure~\ref{fig:vor}).
\begin{figure}
\begin{center}
\includegraphics[width=2.6cm]{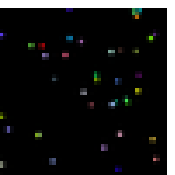}
\includegraphics[width=2.6cm]{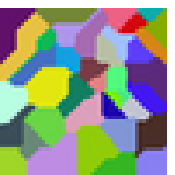}
\includegraphics[width=2.6cm]{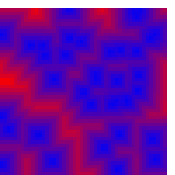}\\
\includegraphics[width=2.6cm]{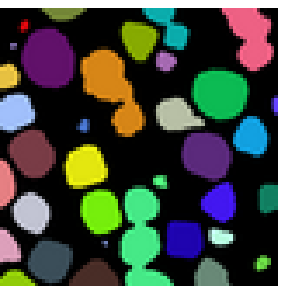}
\includegraphics[width=2.6cm]{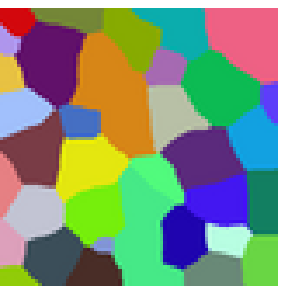}
\includegraphics[width=2.6cm]{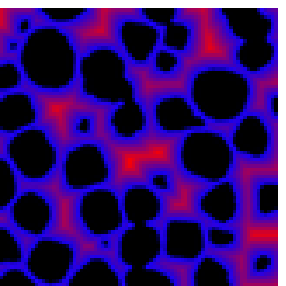}\\
\includegraphics[width=2.6cm]{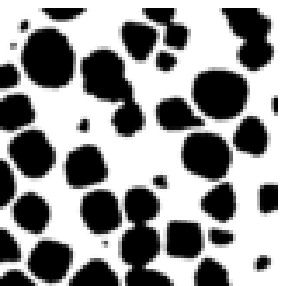}
\includegraphics[width=2.6cm]{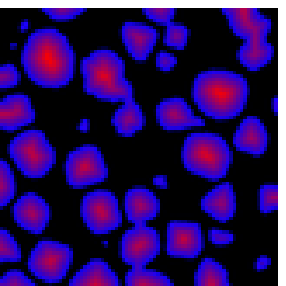}\\
\includegraphics[width=4cm]{image/vicor_perco.eps}
\includegraphics[width=4cm]{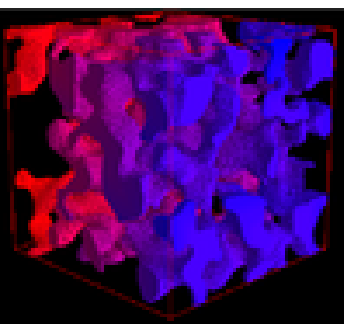}\\
\caption{First serie: the left image is a realisation of a random point process with $\lambda=0.005$ (for the visualisation convenience, the realisation has been dilated), the middle image is the regions at the end of the growing process, the right image is the distance function. Second serie: the same as the first one except the seeds are not some single points but some connected components. Third serie: the same as the first one except the seed is the complementary of the domain. Fourth serie: the first image is the domain and  the second image is the distance function on this domain such as the seed has been localized on the blue face. This distance function is used to calculate the geometrical tortuosity.}
\label{fig:distap}
\end{center} 
\end{figure}

\subsection{The watershed transformation}
\begin{figure}
\begin{center}
\includegraphics[width=4cm]{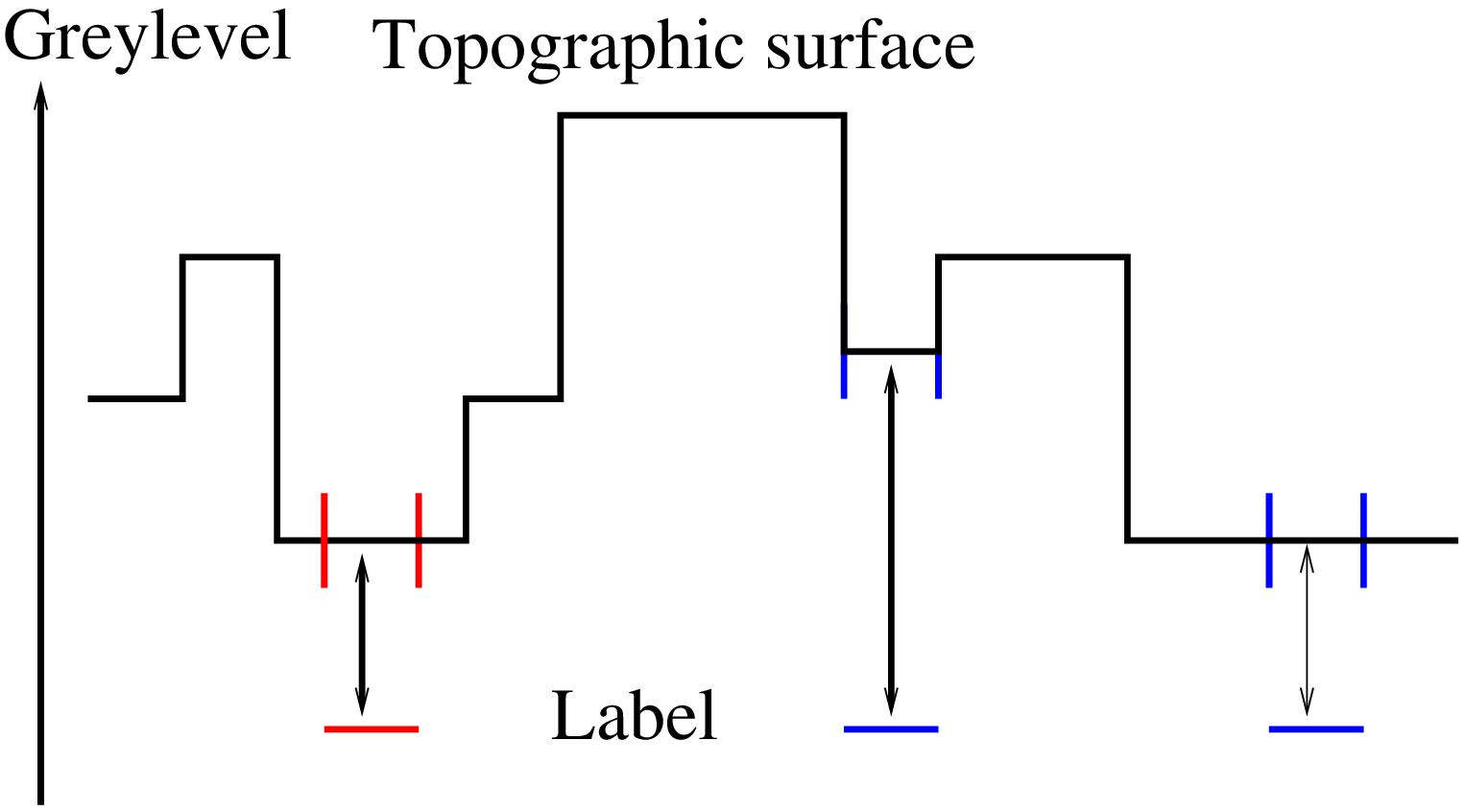}
\includegraphics[width=4cm]{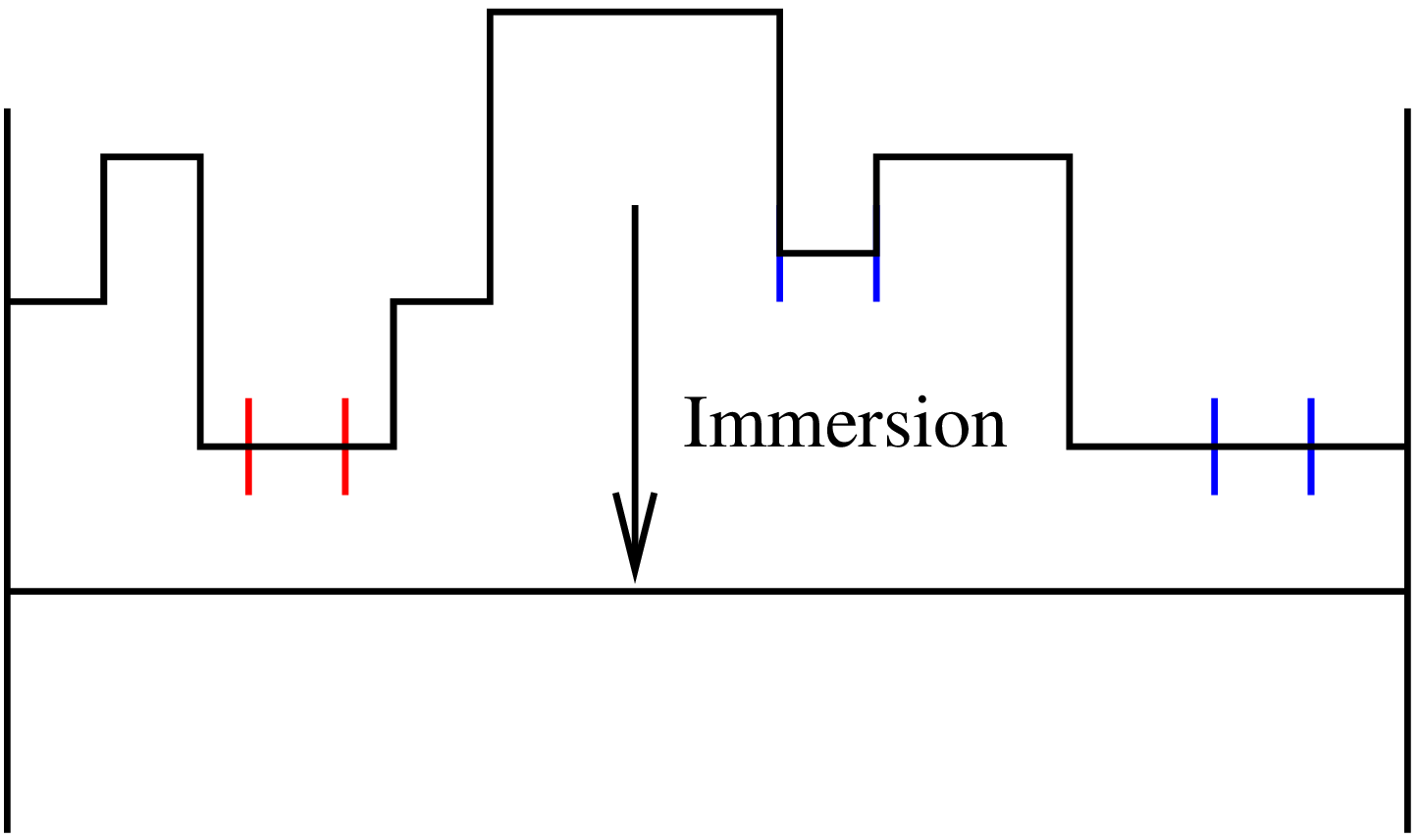}
\includegraphics[width=4cm]{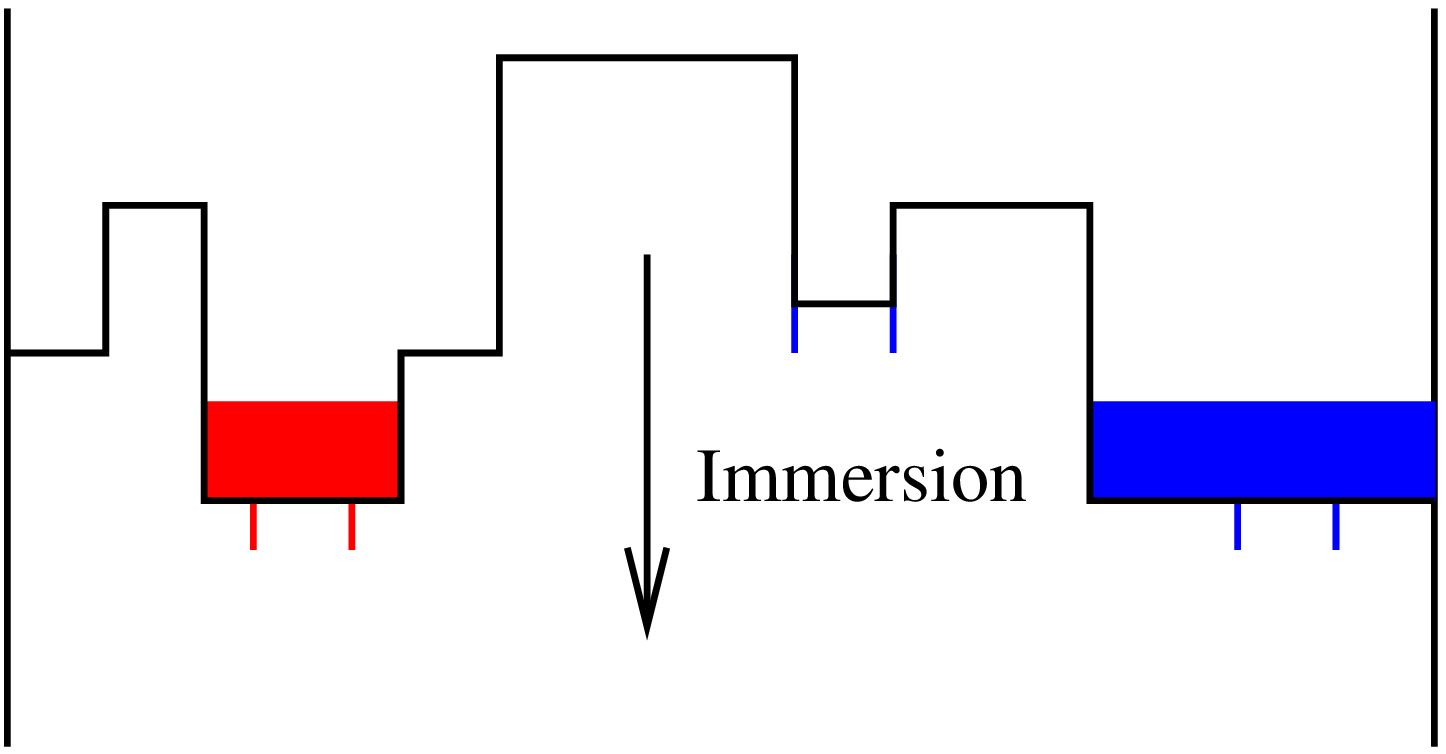}
\includegraphics[width=4cm]{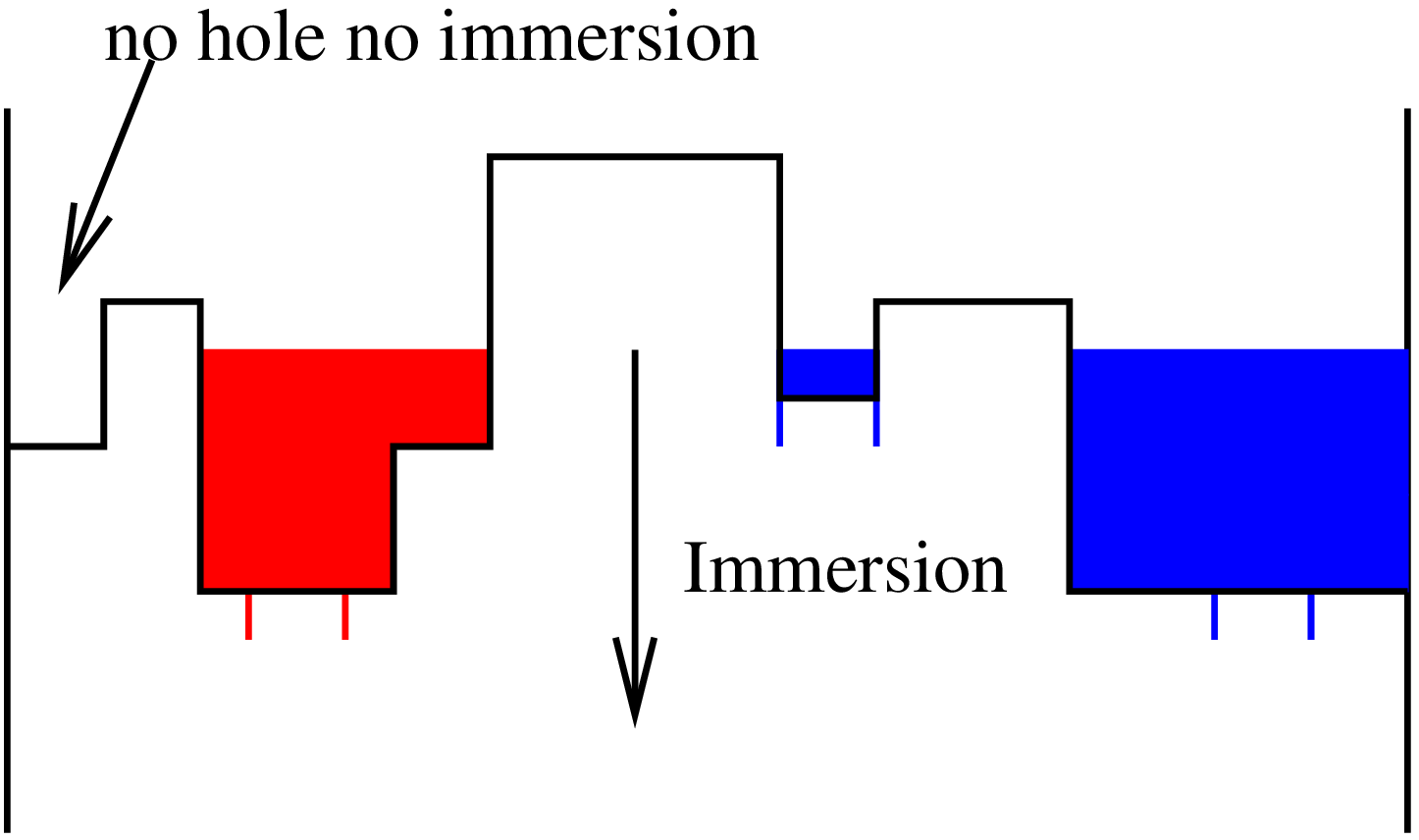}
\includegraphics[width=4cm]{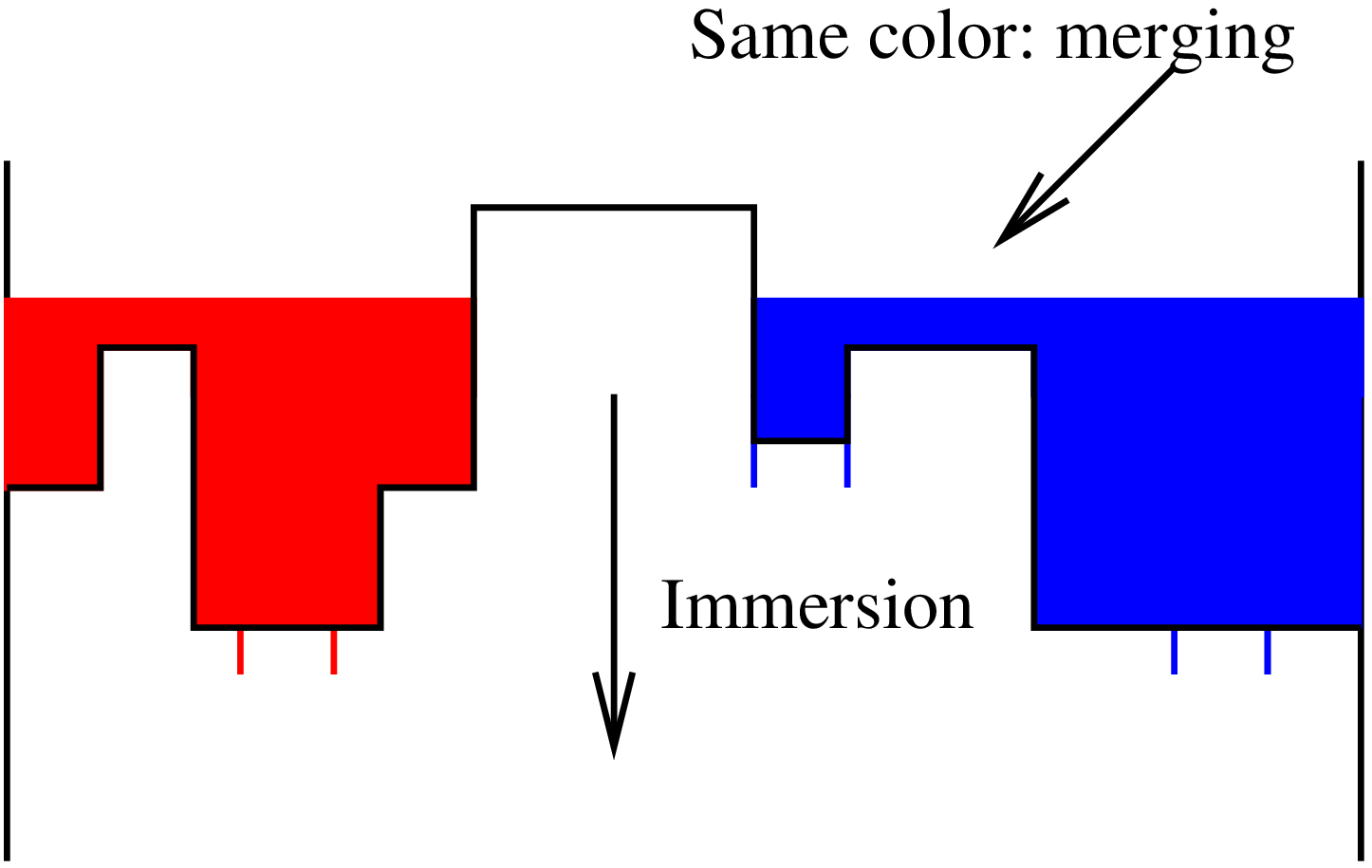}
\includegraphics[width=4cm]{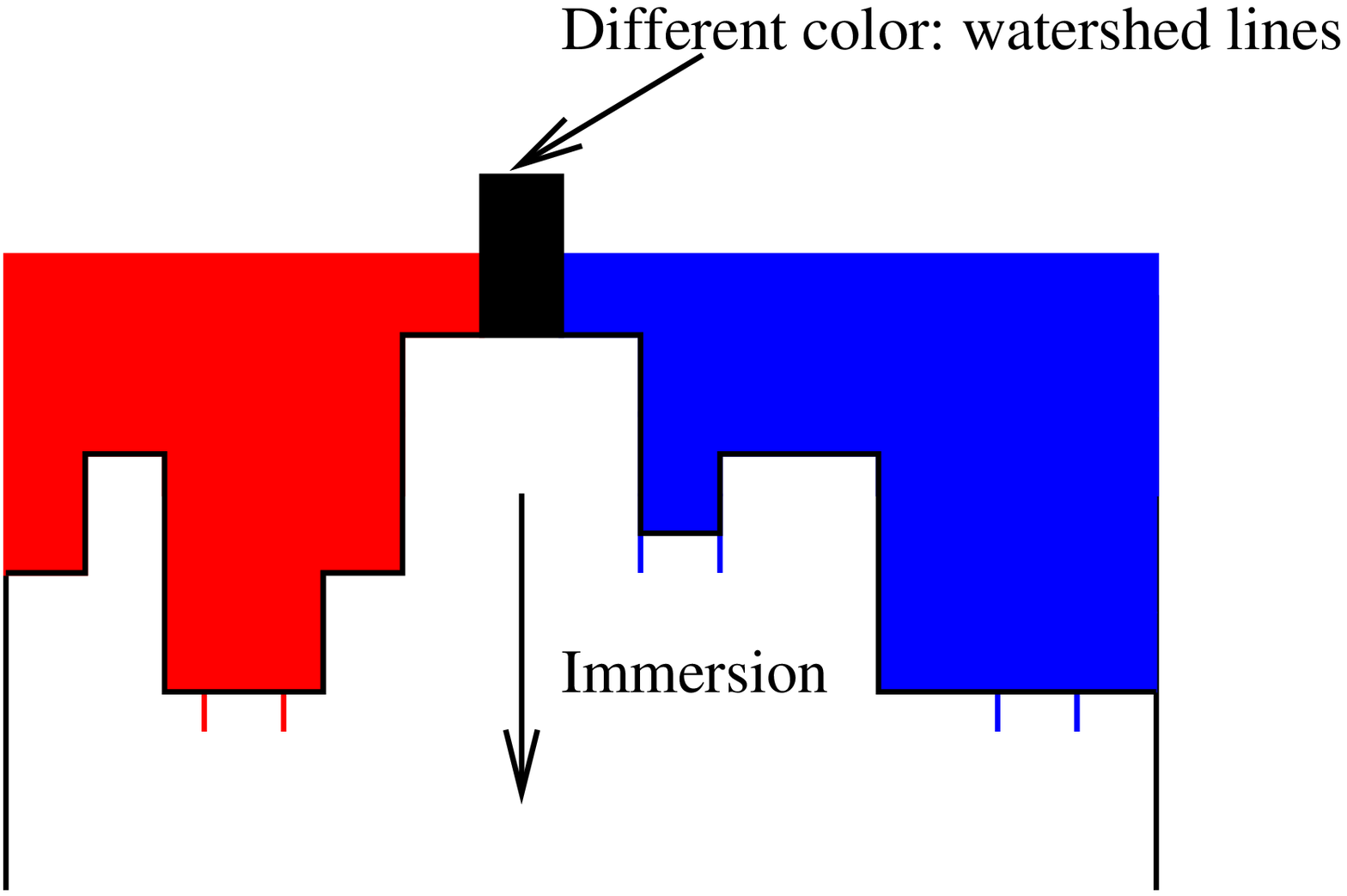}
\caption{This transformation requires two images: the topographic surface (a grey-level image) and the label image (the seeds). The process is: 1) association of each label to a hole 2) immersion 3) the water enters in the topographic by the holes and the catchment basins take the colour of the hole, 4) a part of the topographic is not merged although its level is under the level of the immersion, 5) fusion of two catchment basins with same colour, 6) creation of dam when two catchment basins have different colours. \textit{A video is available at http://pmc.polytechnique.fr/$\sim$vta/water.mpeg}  }
\label{water1}
\end{center} 
\end{figure}

\begin{figure}
\begin{center}
\includegraphics[width=2.6cm]{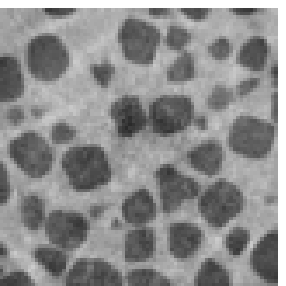}
\includegraphics[width=2.6cm]{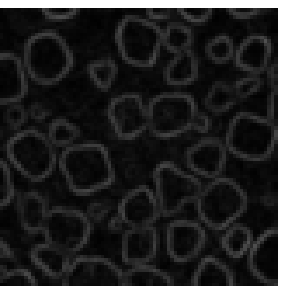}
\includegraphics[width=2.6cm]{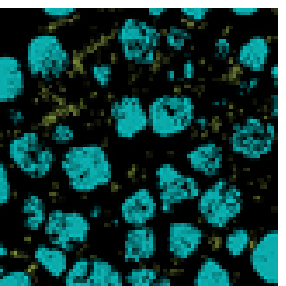}\\
\includegraphics[width=2.6cm]{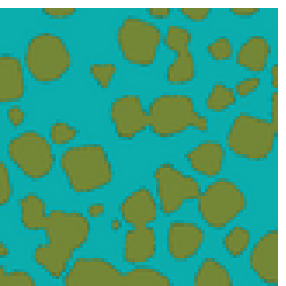}
\includegraphics[width=2.6cm]{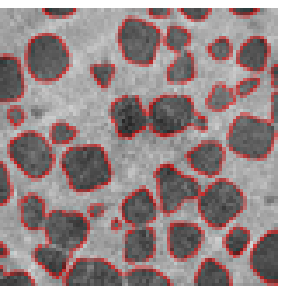}\\
\includegraphics[width=2.6cm]{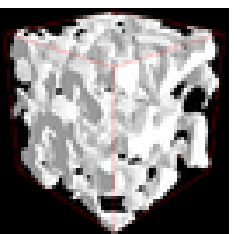}
\includegraphics[width=2.6cm]{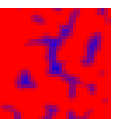}
\includegraphics[width=2.6cm]{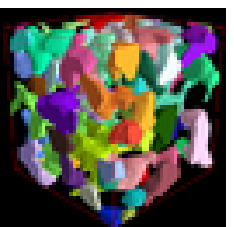}
\caption{The first serie: the first image is the initial image, the second image is the application of a Deriche's gradient\cite{Deriche1987} on the first image, the third image is the visualization of two seeds: one localized on the grains, the other on the grains complementary (in this case, the seeds are not connected). The second serie: the first image is the catchment bassins due to the watershed transformation on the previous gradient image using the two seeds, the second image is the foreground of the boundary region on the initial image. There is a good match with the visual segmentation. The third serie: the first image is the initial image, the second image is a slice of the opposite distance function of the initial image, the third image is the catchment bassins due to the watershed transformation restricted by the initial image on the opposite distance function of the initial image with appropriate seeds. }
\label{water2}
\end{center} 
\end{figure}

An efficient segmentation procedure developed in mathematical morphology is
the watershed segmentation \cite{Beucher1979}, usually implemented by a
flooding process from labels (seeds).\\
Any greyscale image can be considered as a topographic
surface and all boundaries as sharp variations of the grey level. When a
gradient is applied to an image,  boundaries are enhanced. When the
topographic  surface obtained from the gradient is flooded from its seeds,
the waterfronts meet  on  watershed lines in 2D, and on watershed surfaces
in 3D.  A partition of the investigated volume is obtained, where the
catchments basins are separated by the watershed surfaces (see figure~\ref{water1}).\\
%  In the usually application, the seeds are the minima of the gradient image. However, in
% practice, this merging produces an important over-segmentation due to noise
% or local irregularities in the gradient image, generating a set of
% uncontrolled and unwanted markers. To avoid this problem coming from too
% many minima, usually the image, $f$, is filtered. It is a composition of
% vertical (with respect to the grey level) and horizontal filters, in order
% to individualize each grain with a single marker. This individualisation step is difficult\cite{Tariel2008a}. It is the reason why the labels controlled watershed is used. In a next paper \cite{Tariel2008e}, a simple, generic and robust segmentation of granular materials obtained by X-ray tomography will be presented using this algorithm. In this paper, we focus on the implementation.\\
To implement this algorithm, a step by step growing process is defined where a step corresponds to a level of immersion.
Let $level$ be the level of immersion. The ordering attribute function is: $\delta(x,i) = \max(level,f(x))$ such as all the points immerged at the same level are stored in the same queue. The growing process is:
\begin{itemize}	
\item int level=f.min$\_$range();
\item initialization of the regions/ZI by the seeds.
\item for level = f.min$\_$range()  to f.max$\_$range()
\begin{itemize}
\item select the queue number level
\item while the selected queue is not empty
\begin{itemize}
\item extract $(y,i)$ from the selected queue
\item "growth on $x$ of the region $i$"
\end{itemize}
\end{itemize}
\item return regions
\end{itemize}
The quote "growth on $x$ of the region i" means that there are different kinds of growing process introduced in the previous article~\cite{Tariel2008c}. In this example, the growing process is done without a boundary region to divide the other regions. In the algorithm~\ref{alg:water}, the growing process leading up to a final partition invariant about the seeded region initialisation order is used.\\
This growing process is  not limited to the watershed transformation on $E$. The growing process can be restricted to a domain $\Omega=\{\forall x\in E: I(x)\neq 0\}$ if the ordering attribute function is: $\delta(x,i)=\max(level,f(x)) \mbox{ if } I(x)\neq 0, OUT \mbox{ else }$ (see figure~\ref{water2}).
 \begin{algorithm}[h!tp]
\caption{The watershed transformation with an invariant boundary}
\label{alg:water}
\algsetup{indent=1em}
\begin{algorithmic}[20]
 \REQUIRE $f$, $S$ , $V$ \textit{//the topographic image, the seed, the neighbourhood}
\STATE \textbf{\textit{// initialization}}
\STATE int $level$=0;
\STATE System$\_$Queue s$\_$q( $\delta(x,i)=\max (f(x),level) $, FIFO, $f$.max$\_$range() - $f$.min$\_$range()+1); \textit{//n FIFO queues.}
\STATE Population p (s$\_$q); \textit{//create the object Population}
\STATE Restricted $N=\mathbb{N}$;
\STATE Tribe passive($V=\emptyset$);
\STATE int ref$\_$boundary   = p.growth$\_$tribe(passive);
\STATE Tribe active(V, N);
\FORALL{$\forall s_i\in S$ in the order $0,1\ldots$} 
\STATE int ref$\_$tr   = p.growth$\_$tribe(actif); \textit{//create a region/ZI, $(X^t_i,Z^t_i)$ such as $Z^t_i=(X_{i}^t \oplus V)\setminus (\bigcup\limits_{j \in \mathbb{N}} X_{j})$}
\STATE  p.growth($s_i$, ref$\_$tr ); 
\ENDFOR
\STATE \textbf{\textit{//the growing process}}
\FORALL{\textbf{For }$level =$ f.min$\_$range() to f.max$\_$range() }
\STATE  s$\_$q.select$\_$queue($level$); \textit{//Select the queue number level}  
\WHILE{s$\_$q.empty()==false}
\STATE  $(x,i)$=s$\_$q.pop();
\IF{pop.Z()[x].size()$\geq$2 and i= min$\_$elements( pop.Z()[x])}
\STATE  p.growth(x, ref$\_$boundary); \textit{//growth of the boundary region}
\ELSE
\STATE p.growth(x, i ); \textit{//simple growth}
\ENDIF
\ENDWHILE
\ENDFOR
\RETURN pop.X();
\end{algorithmic}
 \end{algorithm}
\subsection{Geodesic reconstruction}
\begin{figure}
\begin{center}
\includegraphics[width=8cm]{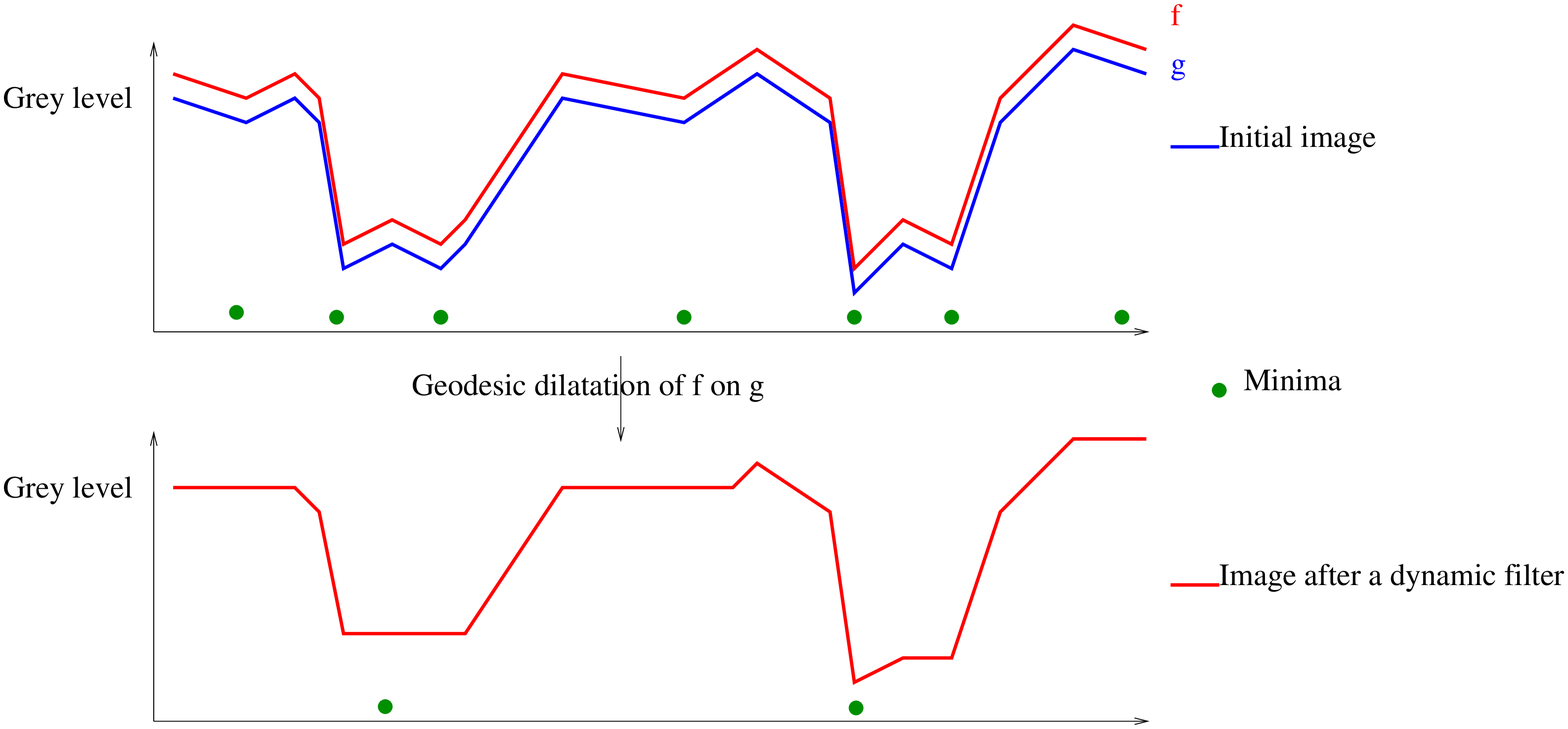}
\caption{The dynamic filter. Before the application of the dynamic filter, there are many minima (green bullets). After the application of the dynamic filter, there are only two minima.}
\label{dynamic}
\end{center} 
\end{figure}

\begin{figure}
\begin{center}
\includegraphics[width=6cm]{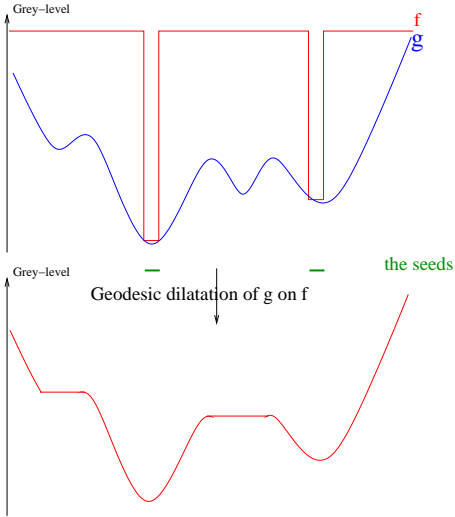}
\caption{The initial image $g$ with seeds. The function $f$ is equal to '$g(x)$ ' if $x$ belongs to the seeds and '$\infty$' if not.}
\label{fig:homo}
\end{center} 
\end{figure}

\begin{figure}
\begin{center}
\includegraphics[height=1.5cm]{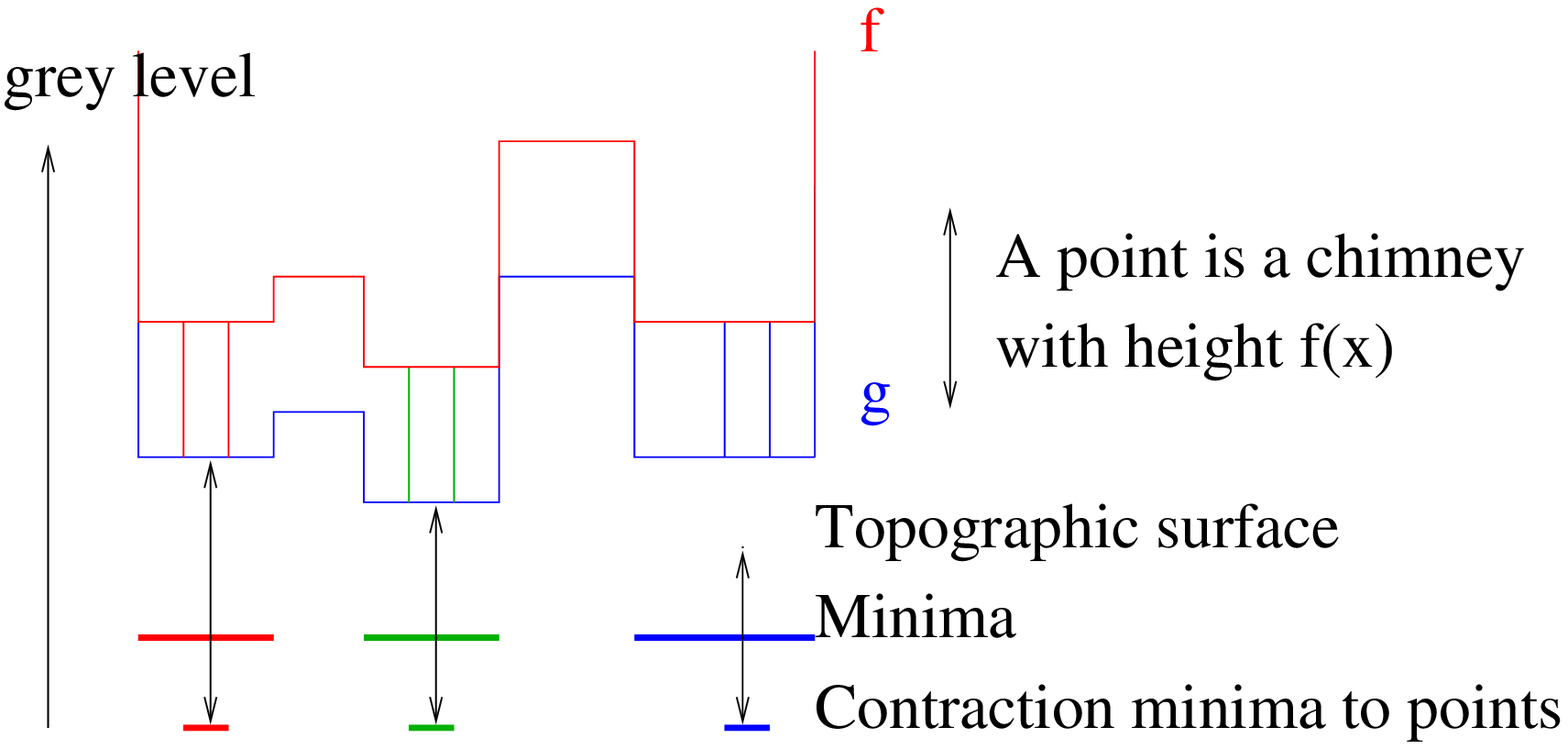}
\includegraphics[height=1.5cm]{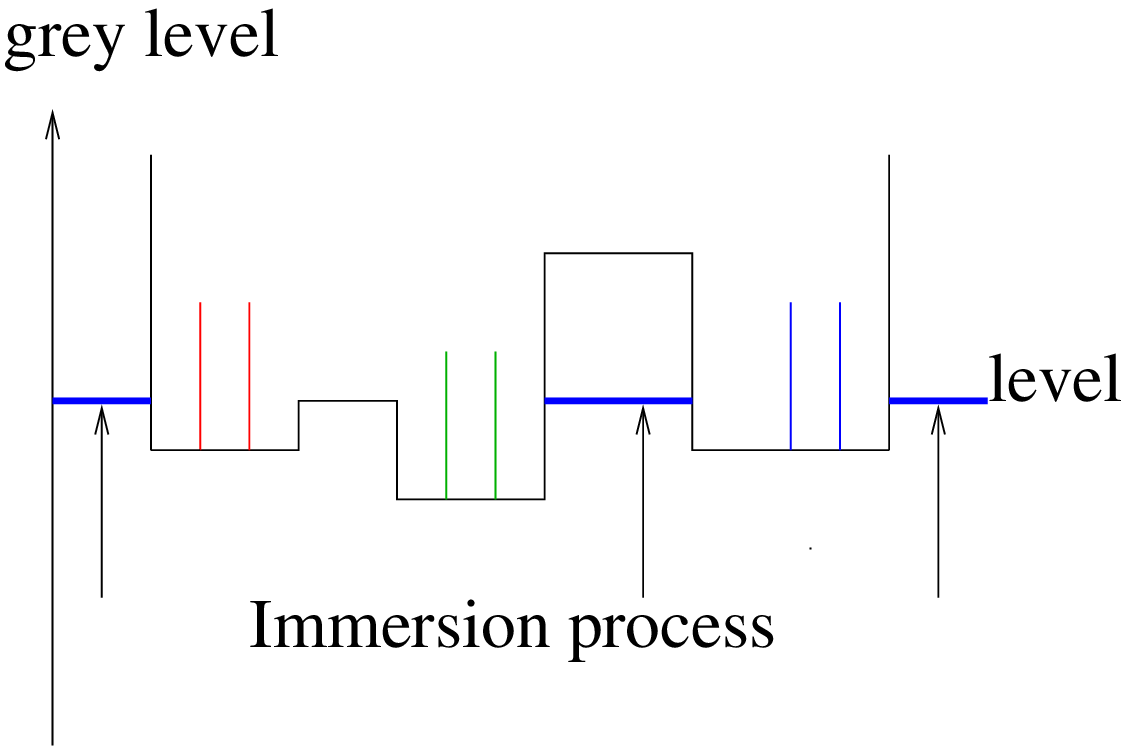}
\includegraphics[height=1.5cm]{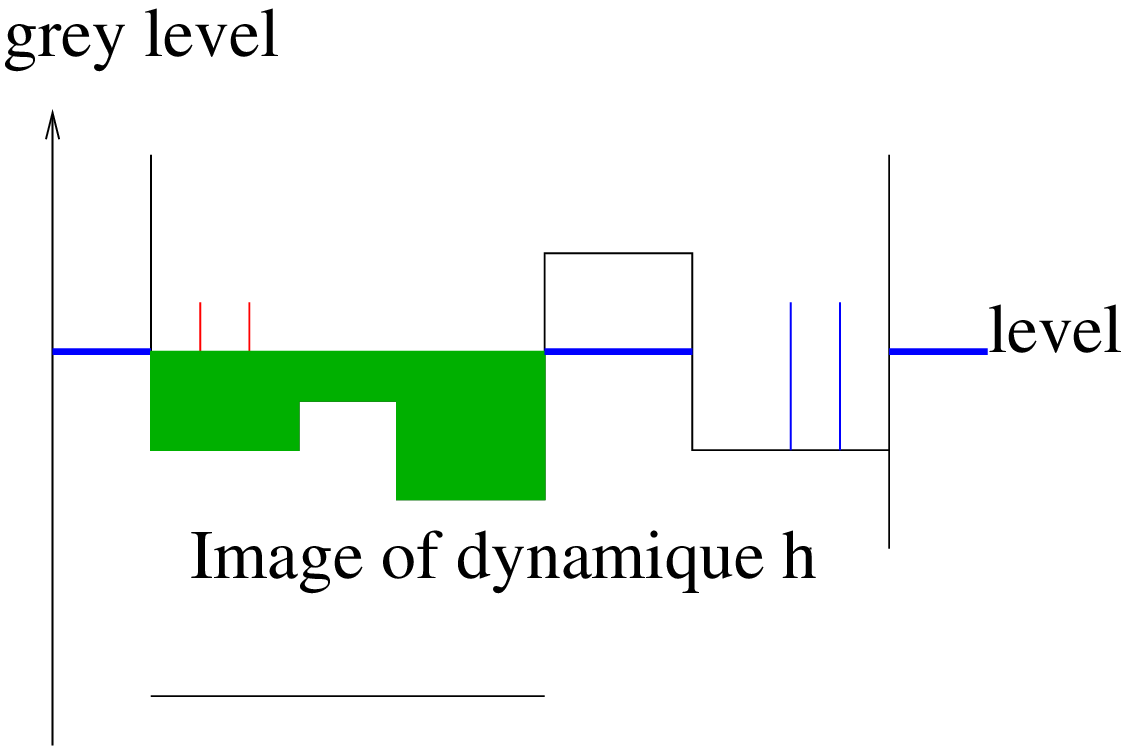}
\includegraphics[height=1.5cm]{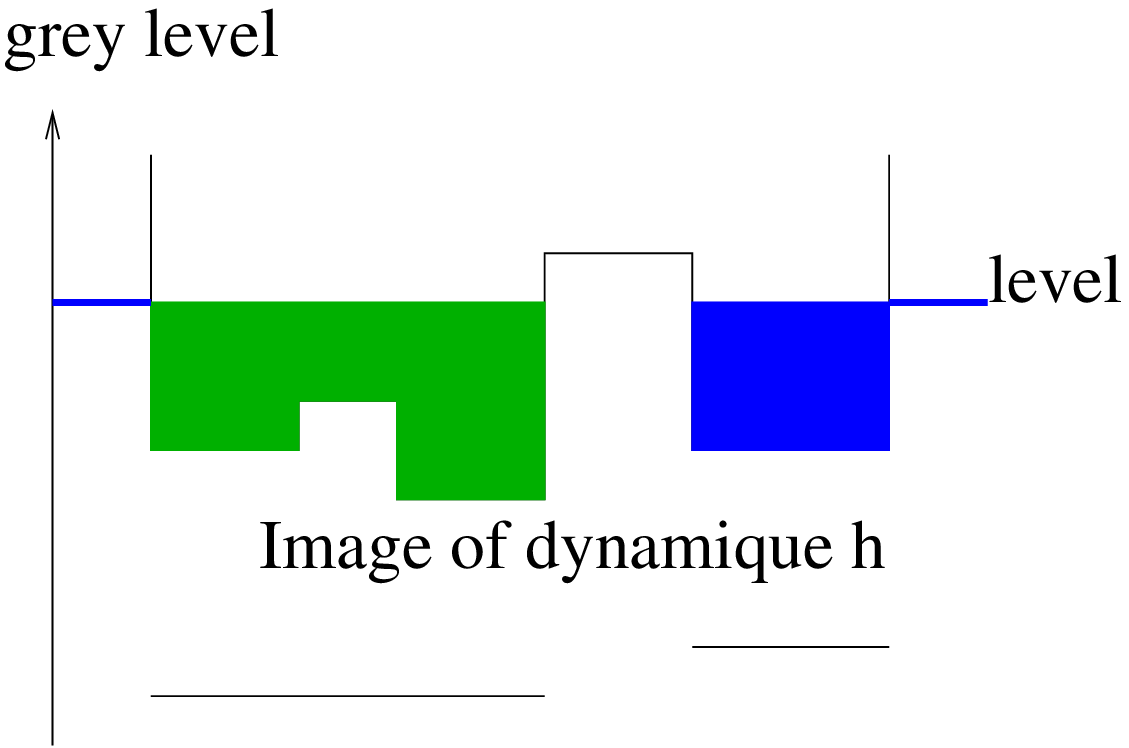}
\includegraphics[height=1.5cm]{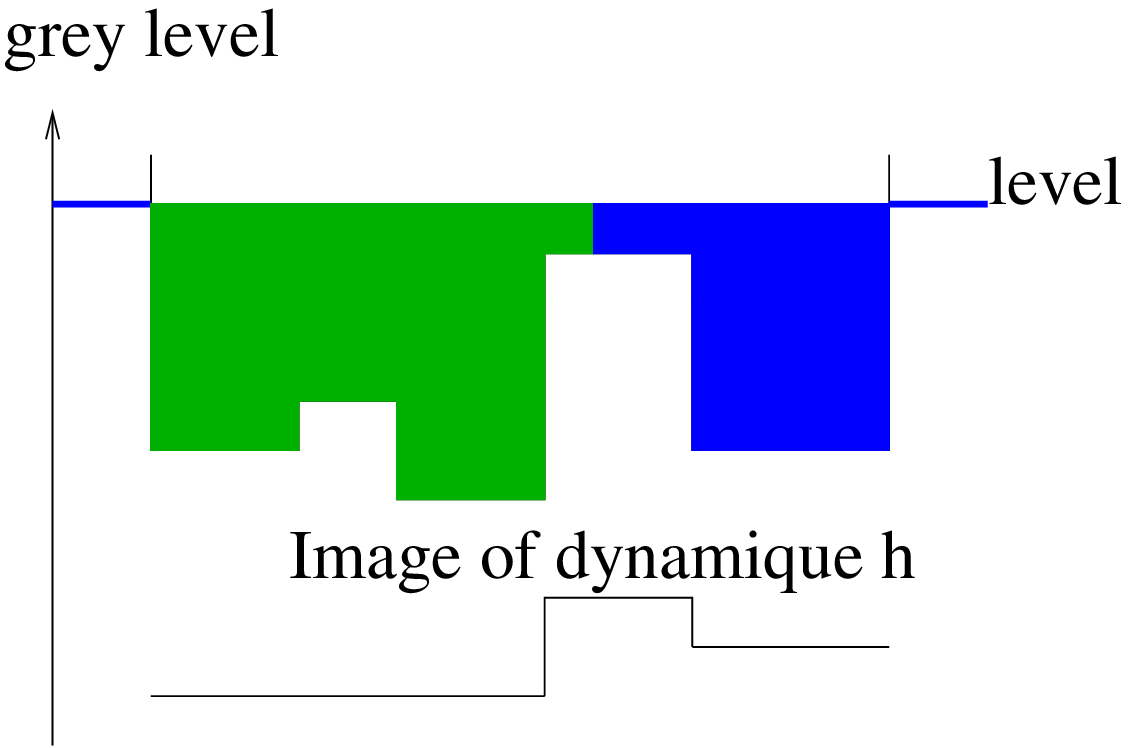}
\caption{The process is: image 1) extraction of the regional minima, contraction of these regional minima to single pixels $x_i$, association between the chimneys with height $f(x_i)-g(x_i)$ and the single pixels; image 2) immersion process: the water enters in the topographic surface by the chimneys if there is not a region yet; image  3) catchment basin takes the colour of the chimney and at every growth $x$ of a region, the  dynamic image takes the immersion level in $x$; image 4) the red chimney does not create region/ZI because the green region is already here. Note that there are 3 minima in the initial image and only two after the dynamic filter.}
\label{principdynamic}
\end{center}
\end{figure}

\begin{figure}
\begin{center}
\includegraphics[width=2cm]{image/domain.eps}
\includegraphics[width=2cm]{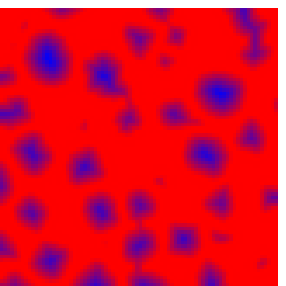}
\includegraphics[width=2cm]{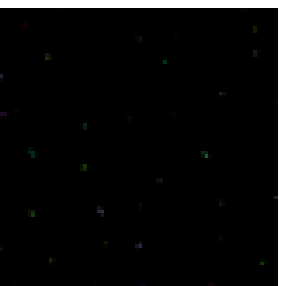}
\includegraphics[width=2cm]{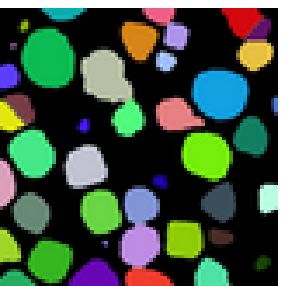}
\includegraphics[width=2cm]{image/domain.eps}
\includegraphics[width=2cm]{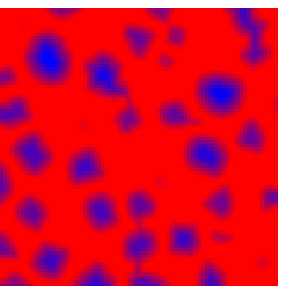}
\includegraphics[width=2cm]{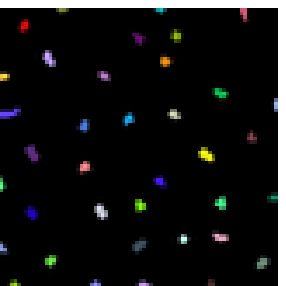}
\includegraphics[width=2cm]{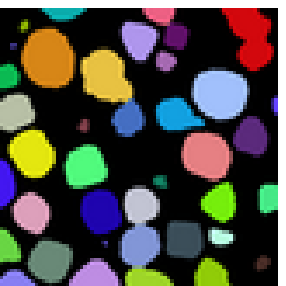}
\includegraphics[width=2cm]{image/domain.eps}
\includegraphics[width=2cm]{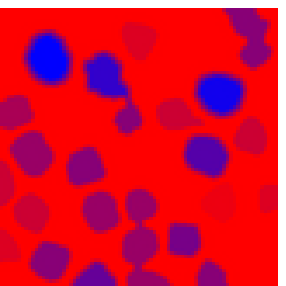}
\includegraphics[width=2cm]{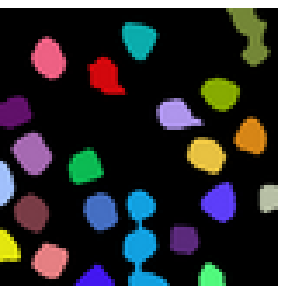}
\includegraphics[width=2cm]{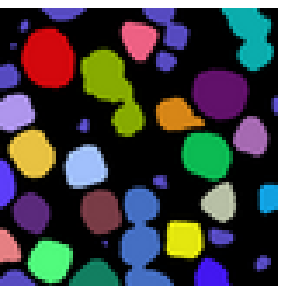}
\includegraphics[width=2cm]{image/vicor_perco2.eps}
\includegraphics[width=2cm]{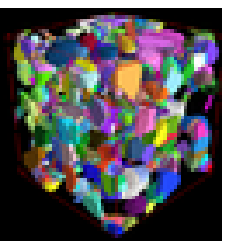}
\includegraphics[width=2cm]{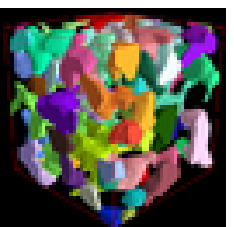}
\includegraphics[width=2cm]{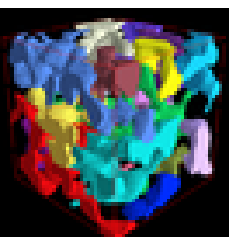}
\caption{For the three first serie, the first image is the initial image, the second image is the opposite of the  distance function  of the initial image after the application of a dynamic filter, (h=0 for the first serie, h=3 for the second, h=10 for the third), the third image is the regional minima of the second image, the fourth image is the watershed tranformation restricted by the first image on the second image using the minima of the third image like a seed. The fourth serie is the same process but in 3D. The first image is the initial image. h=0 for the second image, h=3 for the third, h=10 for the fourth.}
\label{dynav}
\end{center} 
\end{figure}

The geodesic reconstruction is an efficient tool in Morphology Mathematic \cite{Serra1982,Beucher1991}. 
Given a function $f$ and a function $g$ with $f\geq g$, the geodesic erosion is defined as:
\[
R^*_g(f)= E_g^\infty(f)
\]
where $E_g^\infty(f)$ is the infinite geodesic erosion such as $E_g^{t+1}(f)=\sup (E_g^{t}(f)\ominus V,g)$ with $E_g^{0}(f)=f$.\\
Introduced by Grimaud\cite{Grimaud1992}, the geodesic reconstruction is called a dynamic filter when the function $f$ is equal to the function $g$ plus a constant $h$: $f(*)=g(*)+h$. The dynamic filter belongs to the category of vertical filter that fills the valleys with depth lower than $h$ (see figure~\ref{dynamic}).
\\
Introduced by Beucher\cite{Beucher1991}, the geodesic reconstruction is called a homotopic transformation when the function $f$ is equal to $g$ on the seeds, $(s_0,\ldots, s_n)$, and '$\infty$' on the complementary seeds\cite{Beucher2001} (see figure~\ref{fig:homo}). The homotopic transformation is used in the watershed transformation implementation proposed by Vincent \cite{Vincent1991a}  in order to keep only the most significant contours in the areas of interest between the  markers. In our implementation, the homotopic transformation is done during the growing process.\\
The classical implementation of the geodisic reconstruction is to use directly the formula $E_g^{t+1}(f)=\sup (E_g^{t}(f)\ominus V,g)$ with $E_g^{0}(f)=f$. Numerically, the recurrence is stopped when there is nilpotence, $E_g^{t+1}(f)=E_g^{t}(f)$. The implementation is simple but the complexity is  $\Theta (n.k)$, where n is the number of pixels of the image and k is the index of the nilpotence condition.\\
An alternative to this previous algorithm is an algorithm using the SRGPA. The concept of this algorithm is a merging procedure. First, a minima procedure is applied on $f$ to extract the regional minima $(S_{i})_{0\leq i < q}$ of $f$. For the convenience, each $S_{i}$ is reduced to a single pixel $x_i$ thrown randomly in $S_i$. The difference with the watershed transformation is that the creation of region/ZI is done during the merging procedure. At the immersion level equal to $level$, each region/ZI $i$ is created  if $f(x_i)$ is equal to $level$ and if there is not yet a region on the pixel $x_i$. The last difference is that there is not  a region boundary to separate two adjacent regions. At every growth $x$ of a region, the immersion $level$ is attributed  to the dynamic function  on $x$, $E_g^\infty(f)(x)=level$ (see figure~\ref{principdynamic} and algorithm~\ref{alg:dyn}).
The complexity of this algorithm is $\Theta (n)$ where n is the number of pixels of the image. The application is shown on the figure~\ref{dynav}.

 \begin{algorithm}[h!tp]
\caption{Geodesic reconstruction}
\label{alg:dyn}
\algsetup{indent=1em}
\begin{algorithmic}[20]
 \REQUIRE $f$, $g$  $V$, $h$ \textit{//the two images (for the dynamic filter $f=g+h$, the neighbourhood}
\STATE \textbf{\textit{// initialization}}
\STATE Image GR($I$ );
\STATE int $level$=0;
\STATE System$\_$Queue s$\_$q( $\delta(x,i)=\max (g(x),level) $, FIFO,$g$.max$\_$range() - $g$.min$\_$range()+1); \textit{//n FIFO queues.}
\STATE Population p (s$\_$q); \textit{//create the object Population}
\STATE Restricted $N=\mathbb{N}$;
\STATE Tribe active(V, N);
\STATE $(S_{i})_{0\leq i \leq q}$= minima($I$);
\STATE $(x_{i})_{0\leq i \leq q}$= rand$\_$pixel($(S_{i})_{0\leq i < q}$);
\STATE \textbf{\textit{//the growing process}}
\FORALL{\textbf{For }$level =$ I.min$\_$range() to I.max$\_$range() }
\FOR{i=$0$ to q}
\STATE // \textit{\textbf{Creation of region/ZI if two conditions}}
\IF{($level$== $f(x_i))\wedge$pop.X()[$x_i$].empty()==true}
\STATE r$\_$t= p.growth$\_$tribe(actif );
\STATE p.growth($x_i$, r$\_$t );
\STATE GR($x_i$)=level;  
\ENDIF
\ENDFOR
\STATE  s$\_$q.select$\_$queue($level$); \textit{//Select the queue number level}  
\WHILE{s$\_$q.empty()==false}
\STATE  $(x,i)$=s$\_$q.pop();
\STATE p.growth(x, i ); 
\STATE GR(x)=level; 
\ENDWHILE
\ENDFOR
\RETURN GR;
\end{algorithmic}
 \end{algorithm}

\section{Conclusion}
In this paper, we implement various algorithms in the field of SRGPA. Each implementation is simple and efficient using the library Population. When the growing process is done at constant velocity with forgetting the past (simulated Vorono\"i tessellation, domain to clusters, regional minima), a single queue is sufficient to implement these algorithms. When the growing process depends on the topographic surface (watershed transformation and dynamic filter) or when an information has to be keep during the growing process (distance function),  the queues number is more than one  to implement these algorithms.\\
The application of these algorithms will be present in the two next papers of this serie and some new algorithms using the SRGPA will be present in a further paper. 
\appendices
\section{Definition of distance}
\label{ap:dist}
\begin{figure}
\begin{center}
\includegraphics[width=3cm]{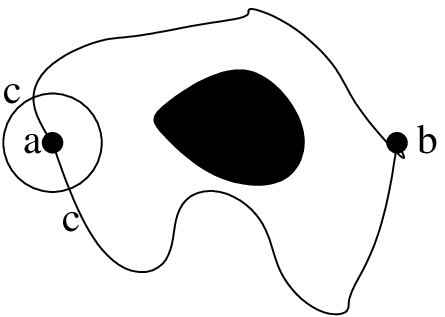}
\includegraphics[width=4cm]{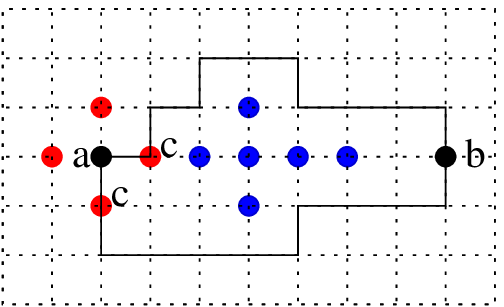}
\caption{Different paths in the metric space for the first image and in the discrete space for the second image. The red bullets are the neighborhood of the point $x$. The blue bullets are the complementary of the domain $\Omega$}
\label{fig:chemin}
\end{center} 
\end{figure}

Let $\Omega$ be a domain of E, a and b two points of $\Omega$. We call
geodesic distance $d_\Omega(a, b)$ in $A$ the lowest bound of the
length of the paths y in $\Omega$, linking a and b.\\
Let $s$ be a set. We call the geodesic distance $d_\Omega(s, b)=\min_{\forall a\in s}d_X(a, b)$, the lowest bound of all geodesic distance $d_\Omega(a, b)$ such as  $a$ belongs to $s$.\\
A property of the geodesic distance is:
\[
d_\Omega(a, b)= \min_{\forall c \in  \overline{\{a\} \oplus V_\epsilon}\cap \Omega }(d_\Omega(a, c) + d_\Omega(c, b))
\]
The symbol $\overline{A}$ means the boundary of $A$.\\
We have especially in the discrete space for the norme $1$ and $\infty$ (see figure~\ref{fig:chemin}):
\[
d_\Omega(a, b)= \min_{\forall c \in  \overline{\{a\} \oplus V_1}\cap \Omega}(1 + d_\Omega(c, b))
\]
\section{Summary of the previous articles}
\label{ap:sum}
The idea of the first article is to define three objects: Zone of Influence (ZI), System of Queues (SQ) and Population. The algorithm implementation using SRGPA is focused on the utilisation of these three objects. An object ZI is associated to each region and localizes a zone on the outer boundary of its region. For example, a ZI can be the outer boundary region excluding  all other regions. An algorithm using SRGPA is not global (no treatment for a block of pixels) but local (the iteration is applied pixel by pixel belonging to the ZI). To manage the pixel by pixel organisation, a SQ sorts out all pixels belonging to ZI depending on the metric and the entering time. It gives the possibility to select a pixel following a value of the metric and a condition of the entering time. The object population links all regions/ZI and permits the (de)growth of regions. A pseudo-library, named Population, implements these three objects. An algorithm can be implemented easier and faster with this library, fitted for SRGPA.\\
The idea of the second article is to give three different growing processes, leading up to three different partitions of the space:
\begin{enumerate}
\item one without a boundary region to divide the other regions,
\item another with a boundary region to divide the other regions,
\item the last one does not depend on the seeded region initialisation order.
\end{enumerate}

\section*{Acknowledgment}
I would like to thank my Ph.d supervisor, P. Levitz, for his support and his trust. The author is indebted to P. Calka for valuable discussion and C. Wiejak for critical reading of the manuscript.  I express my gratitude to the Association Technique de l'Industrie des Liants Hydrauliques (ATILH) for its financial support and the French ANR project "mipomodim" No. ANR-05-BLAN-0017 for their financial support.

\bibliographystyle{plain}
\bibliography {../bibliogenerale}
\end{document}